\documentclass[11pt]{article}

\usepackage[preprint]{acl}

\usepackage{times}
\usepackage{latexsym}

\usepackage[T1]{fontenc}

\usepackage[utf8]{inputenc}

\usepackage{microtype}

\usepackage{inconsolata}

\usepackage{graphicx}
\usepackage{listings}
\usepackage[most]{tcolorbox}
\usepackage{placeins}
\tcbuselibrary{listings}
\title{League of LLMs: A Benchmark-Free Paradigm for Mutual Evaluation of Large Language Models}

\author{
 \textbf{Qianhong Guo\textsuperscript{1,*}},
 \textbf{Wei Xie\textsuperscript{1,*,†}},
 \textbf{Xiaofang Cai\textsuperscript{2}},
 \textbf{Enze Wang\textsuperscript{1}},
 \textbf{Shuoyoucheng Ma\textsuperscript{1}},
 \\
 \textbf{Xiaobing Sun\textsuperscript{3}},
 \textbf{Tian Xia\textsuperscript{1}},
 \textbf{Kai Chen\textsuperscript{2}},
 \textbf{Xiaofeng Wang\textsuperscript{1}},
 \textbf{Baosheng Wang\textsuperscript{1,†}}
\\
 \textsuperscript{1}College of Computer Science and Technology, National University of Defense Technology,
 \\
 \textsuperscript{2}Institute of Information Engineering, Chinese Academy of Sciences,
 \\
 \textsuperscript{3}Institute of High Performance Computing, A*STAR
\\
 \small{
   \textsuperscript{*} Equal Contribution.
   \textsuperscript{†} Correspondence: {\{xiewei, bswang\}@nudt.edu.cn}
 }
 \\
 \small{
   Code is available at \url{https://github.com/Qhovo1/League-of-LLMs}
 }
}

\begin{document}
\maketitle

\begin{abstract}
Although large language models (LLMs) have shown exceptional capabilities across a wide range of tasks, reliable evaluation remains a critical challenge due to data contamination, opaque operation, and subjective preferences. To address these issues, we propose League of LLMs (LOL), a novel benchmark-free evaluation paradigm that organizes multiple LLMs into a self-governed league for multi-round mutual evaluation. LOL integrates four core criteria (dynamic, transparent, objective, and professional) to mitigate key limitations of existing paradigms. Experiments on eight mainstream LLMs in mathematics and programming demonstrate that LOL can effectively distinguish LLM capabilities while maintaining high internal ranking stability (Top-$k$ consistency $= 70.7\%$). Beyond ranking, LOL reveals empirical findings that are difficult for traditional paradigms to capture. For instance, ``memorization-based answering'' behaviors are observed in some models, and higher in-family scores are found in the OpenAI model family ($\Delta = 9$, $p < 0.05$). Finally, we make our framework and code publicly available as a valuable complement to the current LLM evaluation ecosystem.
\end{abstract}

\begin{figure*}[t]
\centering
\includegraphics[width=0.98\textwidth]{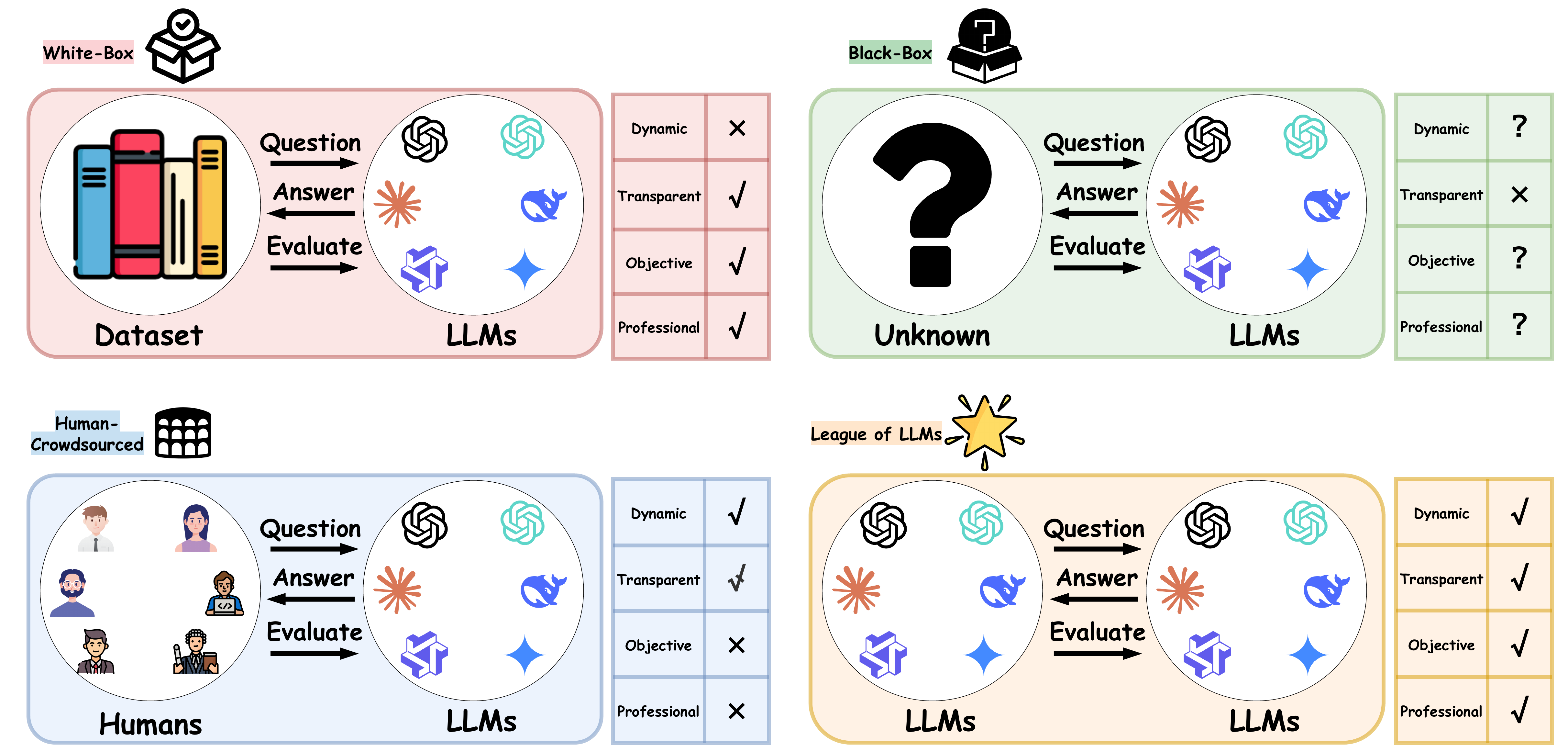} 
  \caption{\textbf{Mainstream LLM evaluation paradigms vs. League of LLMs (LOL).} Compared under four core criteria: Dynamic, Transparent, Objective, and Professional.}
  \label{fig:motivations}
\end{figure*}

\section{Introduction}
Large language models (LLMs) have recently exhibited exceptional capabilities across a wide range of scenarios \cite{achiam2023gpt,team2023gemini,touvron2023llama,ge2023openagi,vaswani2017attention}, including scientific research assistance \cite{boyko2023interdisciplinary}, code generation \cite{nejjar2025llms}, and educational question answering \cite{kasneci2023chatgpt}. As model scale and application scope expand, their capability differences across tasks have become increasingly apparent \cite{chang2024survey}. While they excel at tasks such as mathematical reasoning \cite{ahn2024large,azerbayev2023llemma,luo2023wizardmath,lu2023survey,yu2023metamath} and code generation \cite{xu2022systematic}, their capabilities remain uneven across different domains. Some models perform strongly in logical reasoning, while others are more proficient in engineering implementation or language understanding. These differences not only impact the selection of LLMs for practical applications but also raise the bar for rigorous comparison and development. Consequently, there is an urgent need for a series of systematic, objective, and discriminative evaluation paradigms.

We categorize mainstream LLM evaluation into three paradigms: white-box evaluation, black-box evaluation, and human-crowdsourced evaluation.
While these paradigms exhibit respective advantages in certain aspects, they also suffer from distinct limitations. Most mainstream benchmarks belong to white-box evaluation, which relies on publicly available static datasets, offering strong reproducibility and transparency. 
However, long-term public datasets increase the risk of data contamination: LLMs may memorize test instances through prior training exposure \cite{cheng2025survey,xu2024benchmark,deng2023investigating}, undermining the credibility of evaluation results \cite{sainz2023nlp}.
Black-box evaluation mitigates the risk of data contamination by concealing test datasets, but its opaque operation makes the evaluation difficult to verify and reproduce, limiting credibility \cite{casper2024black}.
Human-crowdsourced evaluation establishes rankings through user preference voting \cite{chiang2024chatbot}. While it reduces leakage risks associated with fixed benchmarks, it heavily relies on users' subjective preferences to ask questions and evaluate answers. Many of these users cannot be assumed to have expert-level domain knowledge \cite{moore2024assessing}, making it difficult to ensure objective and professional evaluation \cite{li2024does}. Together, these limitations motivate a new evaluation paradigm that is dynamic, transparent, objective, and professional.

We introduce League of LLMs (LOL), a novel benchmark-free evaluation paradigm. LOL organizes multiple LLMs into a self-governed league, where they compete for leaderboard positions across multiple rounds. In each round, LLMs take turns (i) generating questions, (ii) answering independently, and (iii) mutually evaluating one another, with the final ranking computed by aggregating the resulting scores.

LOL offers several advantages that existing evaluation paradigms rarely provide simultaneously. By dynamically generating questions, it reduces the overlap between fixed benchmarks and training data, thereby mitigating data contamination. The evaluation process is also designed to be traceable and open to scrutiny, enhancing transparency as well as reproducibility. Furthermore, by aggregating judgments from multiple LLMs rather than relying on a single evaluator or human preference voting, LOL can mitigate the influence of subjective preferences. Both question generation and evaluation are performed by LLMs, which can exhibit more specialized domain knowledge than most crowd users. Thus, LOL can better support more professional and objective evaluation results.

To validate the proposed League of LLMs (LOL) evaluation paradigm, we systematically evaluate eight mainstream LLMs (e.g., gpt-4.1, gemini-2.5-pro-exp, deepseek-r1) across two representative domains: mathematics and programming. The evaluation results indicate that LOL differentiates LLM capabilities while yielding stable rankings across runs (Top-$k$ consistency $= 70.7\%$), supporting the robustness and reliability of the resulting rankings. Beyond ranking, LOL surfaces empirical findings that are rarely captured by traditional paradigms in practice. We observe ``memorization-based answering'' behaviors in some models. Moreover, when LLMs are grouped by developer families, higher in-family scores are found in the OpenAI model family, where in-family scores are higher than out-of-family scores ($\Delta = 9$, $p < 0.05$).

Our contributions are summarized as follows:

\textbf{1. Evaluation Paradigm:} We propose League of LLMs (LOL), a novel benchmark-free evaluation paradigm based on multi-LLM evaluation to mitigate key limitations of existing paradigms.

\textbf{2. Empirical Findings:} We use LOL to systematically evaluate eight LLMs in mathematics and programming, yielding discriminative capability comparisons and revealing empirical findings that traditional paradigms struggle to capture, including ``memorization-based answering'' behaviors and higher in-family scores in the OpenAI model family.

\textbf{3. Community Value:} We publicly release our framework and code to facilitate evaluation of new LLMs and to serve as a valuable complement to the current LLM evaluation ecosystem.

\section{Motivation}

The evaluation of LLMs aims to objectively assess model capabilities and establish rankings, providing guidance for researchers and users. Mainstream LLM evaluation can be categorized into three paradigms: white-box evaluation, black-box evaluation, and human-crowdsourced evaluation.

\subsection{White-Box}
White-box evaluation is a widely used paradigm. It relies on publicly available datasets with reference answers, offering strong transparency and reproducibility. Representative benchmarks include HELM \cite{liang2023holistic}, MATH \cite{hendrycks2021measuring}, GSM8K \cite{cobbe2021training}, HumanEval \cite{chen2021evaluating}, LiveCodeBench \cite{jain2024livecodebench}, and MMLU \cite{hendrycks2020measuring}.

However, since test datasets are fixed and public over long periods, white-box evaluation increasingly suffers from data contamination: LLMs may have encountered test instances during training, undermining evaluation credibility \cite{yang2023rethinking}. Relevant studies also indicate that even minor alterations or rephrasing can significantly reduce the accuracy of LLMs' answers and cause LLMs to reuse the original answers, exhibiting ``memorization'' rather than genuine reasoning \cite{mirzadeh2024gsm,li2024gsm,xie2024large}.

\subsection{Black-Box}
To mitigate data contamination, black-box evaluation uses newly collected or non-public test datasets. For example, Huang \cite{huang2023competition} uses real-time competition data. GPQA \cite{rein2024gpqa} utilizes a newly constructed expert-written question bank. SuperGLUE \cite{wang2019superglue} only returns aggregate scores via a closed server.

Although black-box evaluation improves confidentiality, its opaque operation limits verifiability and auditability. Since the test data and evaluation process are hidden, the evaluation results are difficult to verify by third parties, which weakens credibility, interpretability, and reproducibility \cite{casper2024black}. Furthermore, black-box datasets may still originate from publicly available internet data or be leaked. As long as the datasets remain fixed, the risk of data contamination persists.

\subsection{Human-Crowdsourced}
To reduce data contamination from fixed test datasets, Chatbot Arena \cite{chiang2024chatbot} is proposed as a human-preference-based crowdsourced evaluation platform. This platform adopts pairwise comparisons: two LLMs answer the same user question, users select the better answer, and the system aggregates pairwise outcomes to produce an overall ranking. 

Despite its inherent dynamism, this paradigm has several limitations. First, it relies heavily on users' subjective preferences (e.g., longer or containing emojis tend to be favored \cite{sentimentarena2025,li2024does}). Moreover, user-generated questions vary widely in quality, and most users cannot be assumed to have expert-level domain knowledge. Finally, although Chatbot Arena discloses its ranking rules and metric computation methods, its core raw data (user inputs, LLM outputs, and evaluation records) remain non-public, leaving key parts of the evaluation process opaque and difficult to independently verify or audit.

\subsection{Evaluation Criteria}
In summary, we argue that scientific LLM evaluation should satisfy four core criteria jointly.

\textbf{1. Dynamic:} Each evaluation should use questions and answers with sufficient freshness and diversity to mitigate data contamination.

\textbf{2. Transparent:} Evaluation should be open and transparent, with reproducible processes and results, thereby enhancing credibility.

\textbf{3. Objective:} Evaluation standards and results should not be influenced by individual subjective preferences, thereby reducing bias and the possibility of manipulation, and reflecting the objective performance of LLMs.

\textbf{4. Professional:} Questions and answers should be at or near the average level of human experts, enabling in-depth evaluations of LLMs' professional capabilities in vertical domains.

\begin{figure*}[t]
\centering
\includegraphics[width=0.98\textwidth]{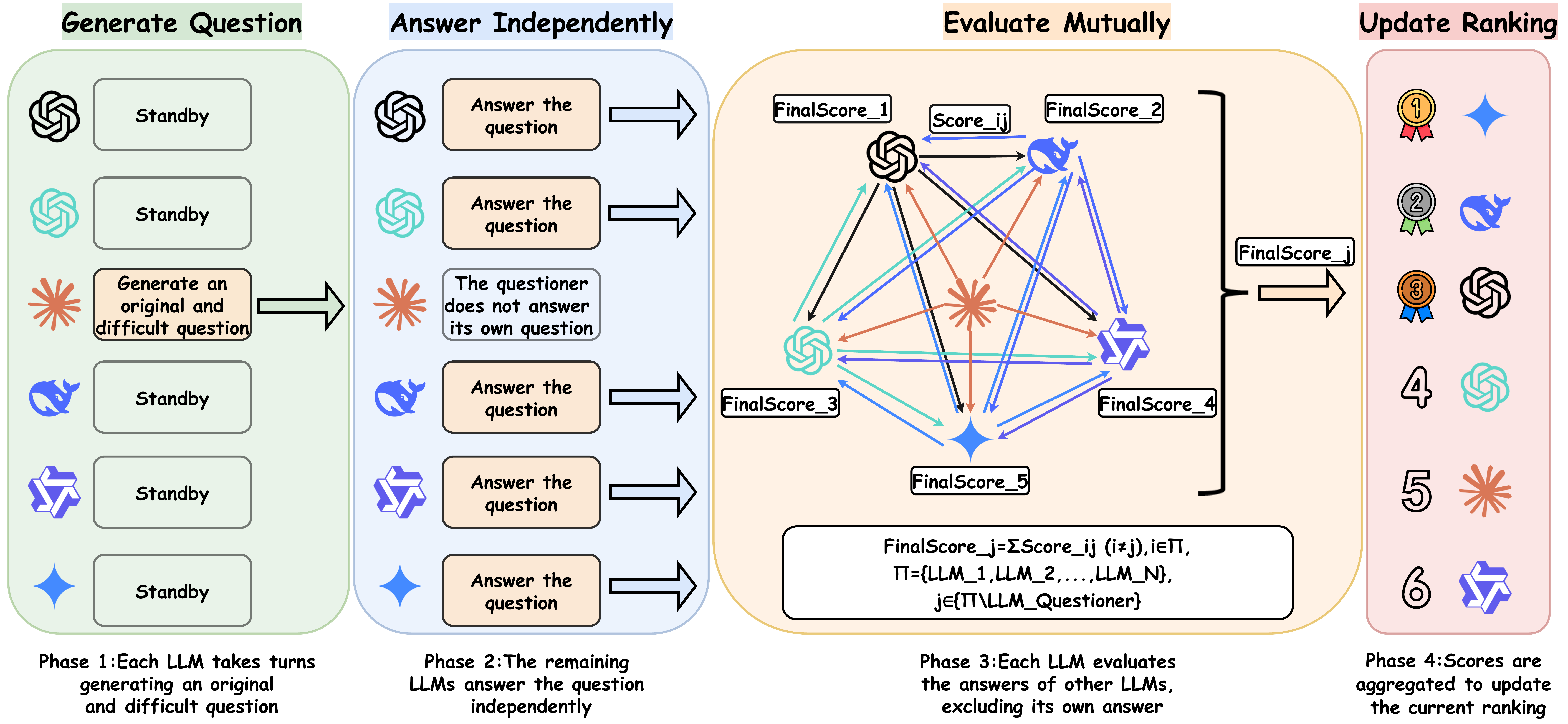}
  \caption{\textbf{Overview of the League of LLMs evaluation pipeline.} It consists of four phases: Generate Question, Answer Independently, Evaluate Mutually, and Update Ranking.}
  
  \label{fig:methodology}
\end{figure*}

\section{League of LLMs}

As shown in Figure \ref{fig:motivations}, white-box evaluation (based on public datasets) suffers from insufficient dynamism, while black-box evaluation (based on unknown datasets) lacks transparency. Human-crowdsourced evaluation (based on human evaluators) is influenced by subjective preferences, incomplete transparency, and limited professionalism. To mitigate these limitations, we propose LOL, a benchmark-free evaluation paradigm in which multiple LLMs generate questions, answer independently, and evaluate mutually, with final rankings computed by aggregating scores across rounds.

\subsection{Evaluation Pipeline}
As illustrated in Figure \ref{fig:methodology}, our paradigm comprises four phases. Before the evaluation begins, it is necessary to predefine the evaluation domain and rules to ensure controllability and consistency in both question generation and evaluation processes.

\textbf{Phase 1: Generate Question} 
Each LLM takes turns acting as the questioner, generating an original and challenging question according to the predefined rules, and providing a reference answer. Subsequently, this question is distributed to the other LLMs for answering. 

\textbf{Phase 2: Answer Independently} 
Except for the questioner, the remaining LLMs independently answer the question without referencing others' answers, thereby evaluating their independent question-solving abilities.

\textbf{Phase 3: Evaluate Mutually} 
The questioner LLM provides the reference answer for the other LLMs. Based on the reference answer and the predefined scoring criteria, each LLM evaluates the answers of other LLMs (excluding its own), enabling mutual evaluation among the LLMs.

\textbf{Phase 4: Update Ranking} 
The mutual evaluation results for the current round are aggregated to update the real-time ranking. This ranking serves as periodic feedback and provides the basis for subsequent evaluation. After n rounds, the final ranking is output.

\subsection{Advantages}
The proposed League of LLMs (LOL) evaluation paradigm integrates the strengths of existing paradigms while mitigating their inherent limitations, aiming to establish a more scientific and standardized evaluation paradigm.

\textbf{In terms of dynamism}, evaluation questions are dynamically generated by LLMs based on predefined rules, and evaluation is conducted under predefined scoring criteria. This mechanism mitigates data contamination over time, thereby improving the long-term validity of the evaluation.

\textbf{In terms of transparency}, the processes of question generation, answering, and evaluation are all fully open and traceable. External researchers can independently verify the results, thereby enhancing the interpretability and credibility of the evaluation.

\textbf{In terms of objectivity}, LOL employs a decentralized mutual evaluation mechanism, in which multiple LLMs jointly perform the evaluation task. This reduces the influence of any single evaluator's bias and human subjective preferences \cite{chen2024humans,zhang2025lists}, enhancing the fairness and stability of the evaluation outcomes.

\textbf{In terms of professionalism}, compared to most crowd users, LLMs can exhibit more professional domain knowledge, enabling them to deliver more specialized and higher-quality questions, answers, and scores in most evaluation tasks.

\subsection{Overview of Experiment Setups}
The LOL evaluation paradigm is highly extensible. Its components, such as LLM selection, task domains, scoring mechanisms, and evaluation pipeline, can be flexibly replaced to accommodate different task requirements. Moreover, using eight questions per round is only a lightweight experimental setting. It can be easily scaled by increasing the number of questions or rounds, without human scoring or adjudication.

For instance, in this paper, we validate LOL in two representative domains: mathematics and programming. Within a unified evaluation paradigm, mathematics tasks adopt the Borda rule \cite{borda1781memoire}, while programming tasks employ absolute scores (details in Section \ref{sec:Math_ED} and \ref{sec:Programming_ED}), thereby demonstrating the paradigm's generality.

Considering that LLMs may exhibit biases such as self-preference or conformity during evaluation \cite{liu2023llms,ye2024justice}, we introduce three mechanisms: independent contexts, prohibition of self-evaluation, and multi-LLM mutual evaluation based on reference answers, which further enhance fairness and robustness.

All main experiments were conducted using Huiyan's unified API \cite{huiyan_api}, with temperature set to 1 and all other parameters set to their defaults. We selected eight mainstream LLMs ($n = 8$) as follows: gpt-4.1-2025-04-14, o3-mini, o1, claude-3-7-sonnet-20250219, deepseek-r1, deepseek-v3-0324, qwen2.5-max, and gemini-2.5-pro-exp-03-25. For each domain, we ran 5 independent experimental sets. Each set contained $8$ questions $\times 7$ answers $\times 7$ evaluations, totaling 1,960 data points per domain, comparable in scale to mainstream benchmarks such as MATH500.

\section{League of LLMs On Mathematics}

\subsection{Experiment Design}
\label{sec:Math_ED}
\textbf{Question Generation Mechanism:} To achieve in-depth evaluation and to mitigate data contamination caused by static question banks, we design a professional and dynamic question generation phase (prompts are provided in Appendix \ref{sec:Math Prompts}). The questioner LLM is tasked with acting as a mathematician, identifying the most difficult questions based on its prior knowledge, abstracting the core mathematical principles, and constructing an original and challenging question with a reference answer.

\textbf{Evaluation Mechanism:} Given that mathematical questions often admit multiple valid solution paths, we adopt the Borda rule to reduce subjectivity in absolute scoring. This rule maps the rank order of answers to discrete scores (1st: 6 points, 2nd: 5 points, ..., 7th: 0 points), thereby enabling a more stable and objective comparison.

\begin{figure}[t]
\centering
\includegraphics[width=0.98\columnwidth]{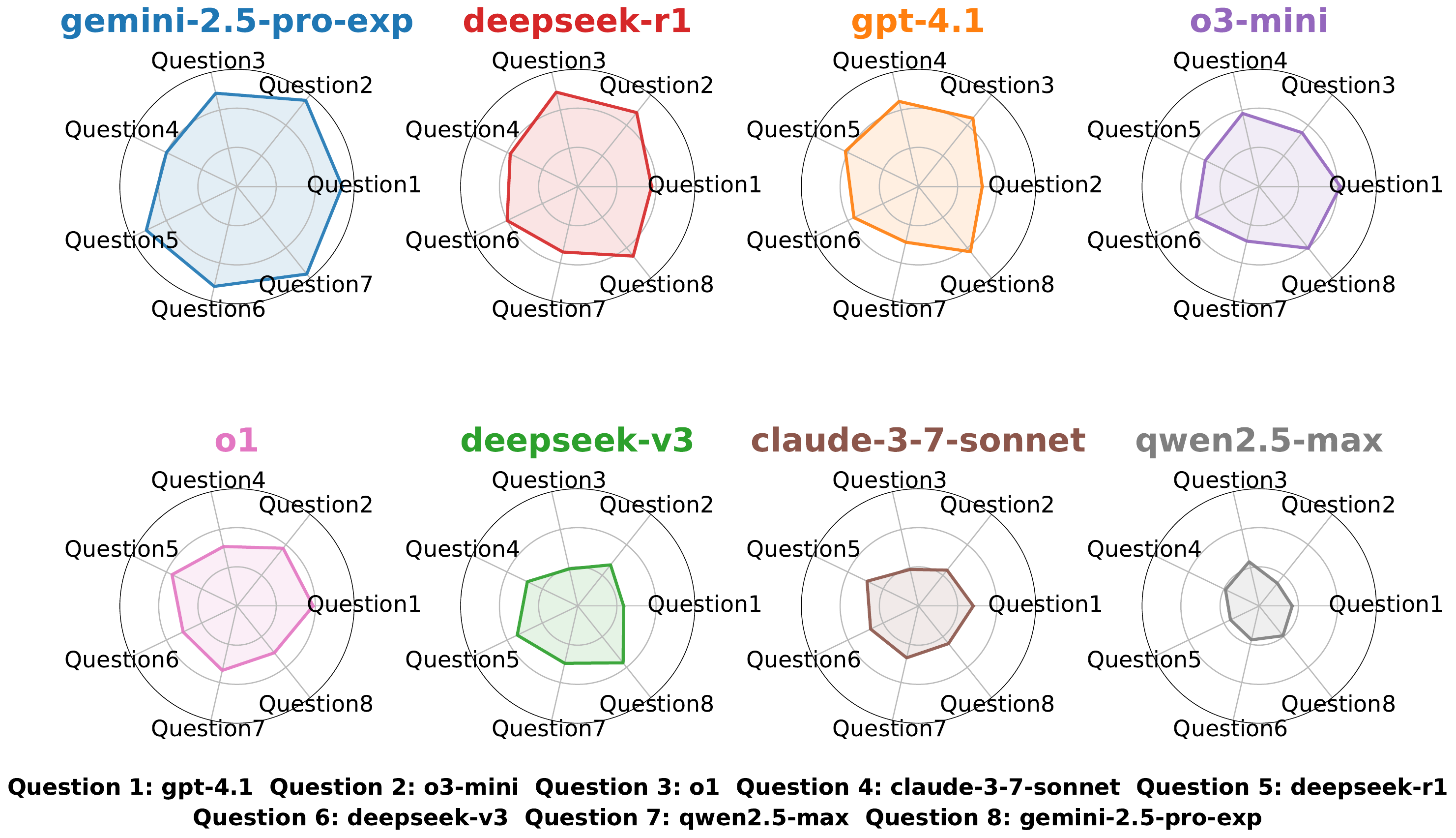} 
\caption{\textbf{Radar charts of LLM performance on mathematics tasks.} They display the mean Borda scores of eight LLMs on eight questions, enabling comparison of their overall performance.}

\label{fig:radar}
\end{figure}

\subsection{Experiment Results}
Figure \ref{fig:radar} illustrates each LLM's scores on questions, reflecting their overall performance. Visually, gemini-2.5-pro-exp covers the largest area, whereas qwen2.5-max covers the smallest. 

To quantify this visual difference, a one-way ANOVA reveals significant differences among the LLMs ($F(7,1952) = 118.71$, $p < 0.001$). Tukey post-hoc analysis further shows that gemini-2.5-pro-exp achieves the highest mean score and differs significantly from all other models ($p < 0.001$), whereas qwen2.5-max scores are significantly lower than all other models ($p < 0.001$). In addition, only four model pairs exhibit no significant differences (deepseek-r1 and gpt-4.1, gpt-4.1 and o3-mini, o3-mini and o1, deepseek-v3 and claude-3-7-sonnet). Except for these four pairs, the remaining comparisons are significant, indicating that LOL distinguishes mathematical capabilities among most models. Statistical details are provided in Appendix \ref{sec:Details for Mathematics}.

\textbf{Finding 1: LLMs exhibit substantial variation in mathematical question generation ability, with some reaching professional-level performance.} 

We observe significant differences among LLMs in their ability to generate mathematical questions. Some LLMs, such as gemini-2.5-pro-exp, demonstrate professional-level question generation ability, as shown below:

\setlength{\fboxsep}{6pt}
\setlength{\fboxrule}{0.5pt}

\noindent\fbox{%
  \parbox{\dimexpr\linewidth-2\fboxsep-2\fboxrule\relax}{%
  \textbf{Example question} \par\smallskip
    Let $S_3(n)$ denote the sum of the digits of a positive integer $n$ when written in base $3$. Define a sequence $a_n = (-1)^{S_3(n)}$, for $n \geq 1$. Consider the Dirichlet series
\(
F(s) = \sum_{n=1}^{\infty} \frac{a_n}{n^s}
\)
, defined for complex numbers $s$ where the series converges. Determine if $F(s)$ can be analytically continued to a region including $s=0$, and if so, find the value $F(0)$.
  }%
}

The generated question integrates elements from number systems, automatic sequences, and analytic continuation in complex analysis, exhibiting a high degree of novelty for benchmark-based evaluation. While its construction draws on classic Dirichlet-series constructions in modern number theory (e.g., the Riemann zeta function), the specific combination and instantiation are unlikely to appear in typical public benchmark datasets or human-crowdsourced evaluations. More details can be found in Appendix \ref{sec:Case of Finding 1}. Although most LLMs failed to provide the correct answer, the resulting performance differentiation instead provides evidence for the effectiveness of the ``prior-knowledge-based question generation mechanism'' described in Section \ref{sec:Math_ED}. This mechanism not only distinguishes capability differences among LLMs but also highlights the advantages of LOL in terms of professionalism and dynamism.

\textbf{Finding 2: LLMs exhibit ``memorization-based answering'' behavior rather than genuine reasoning in some cases.} 

We observe that some LLMs do not perform genuine reasoning when answering questions. Instead, they rely on memorization for template matching. We define this phenomenon as ``memorization-based answering'': the LLM mistakenly treats the current question as a classic question with a similar template but different details and directly applies memorized answers or conclusions. 

For instance, in the Dirichlet series question above, deepseek-r1 attempted to formalize $F(s)$ as an infinite product structure similar to the Euler product, thereby producing an incorrect result. This technique is commonly applicable to Dirichlet series with a multiplicative structure, such as the Riemann zeta function, but it does not apply to the current question. See Appendix \ref{sec:Case of Finding 2} for details.

Similar mistakes appeared 22 times among the 280 responses (7.9\%) in experiments, highlighting a limitation of white-box (static benchmark) evaluation: LLMs may rely on template matching rather than genuine reasoning in these cases. In contrast, LOL can identify such behaviors and reveal the true boundaries of the LLM's reasoning abilities.

\textbf{Finding 3: Evaluation results exhibit high internal ranking stability.}

When $k = n/2$, Top-$k$ consistency reaches $70.7\%$ (95\% CI $[0.700, 0.713]$), suggesting stable internal rankings and a cross-LLM consensus judgment. Meanwhile, the standard deviations of the overall scores across runs remain low on a 100-point scale (3.28 to 10.76). Together, these results quantitatively demonstrate that, even without human scoring or adjudication, mutual evaluation among LLMs can converge to stable and consistent results. More details are provided in Appendix \ref{sec:Top-k}.

\section{League of LLMs On Programming}

\subsection{Experiment Design}
\label{sec:Programming_ED}
\textbf{Question Generation Mechanism:} To mitigate the limitation that existing benchmarks emphasize code correctness but overlook question-setting ability, we design a professional and dynamic question generation phase (prompts are provided in Appendix \ref{sec:Programming Prompts}). The questioner LLM acts as a senior programming competition question setter, designing professional-level, ACM-style original programming questions. Each question includes a description, input/output format, constraints, a reference answer, and a complexity analysis. Answering LLMs can freely choose the programming language, avoiding bias introduced by restrictions.

\textbf{Evaluation Mechanism:} To better reflect professionalism and objectivity, we adopt a 100-point scoring scheme. Each LLM evaluates others' answers based on the reference answer and scoring criteria across multiple dimensions, including correctness, efficiency, readability, and modularity.

\subsection{Experiment Results}
Beyond the aggregate scores, LOL reveals two additional findings through mutual evaluation.

\textbf{Finding 4: Question-solving and question-setting abilities are decoupled.}

\begin{figure}[t]
\centering
\includegraphics[width=0.98\columnwidth]{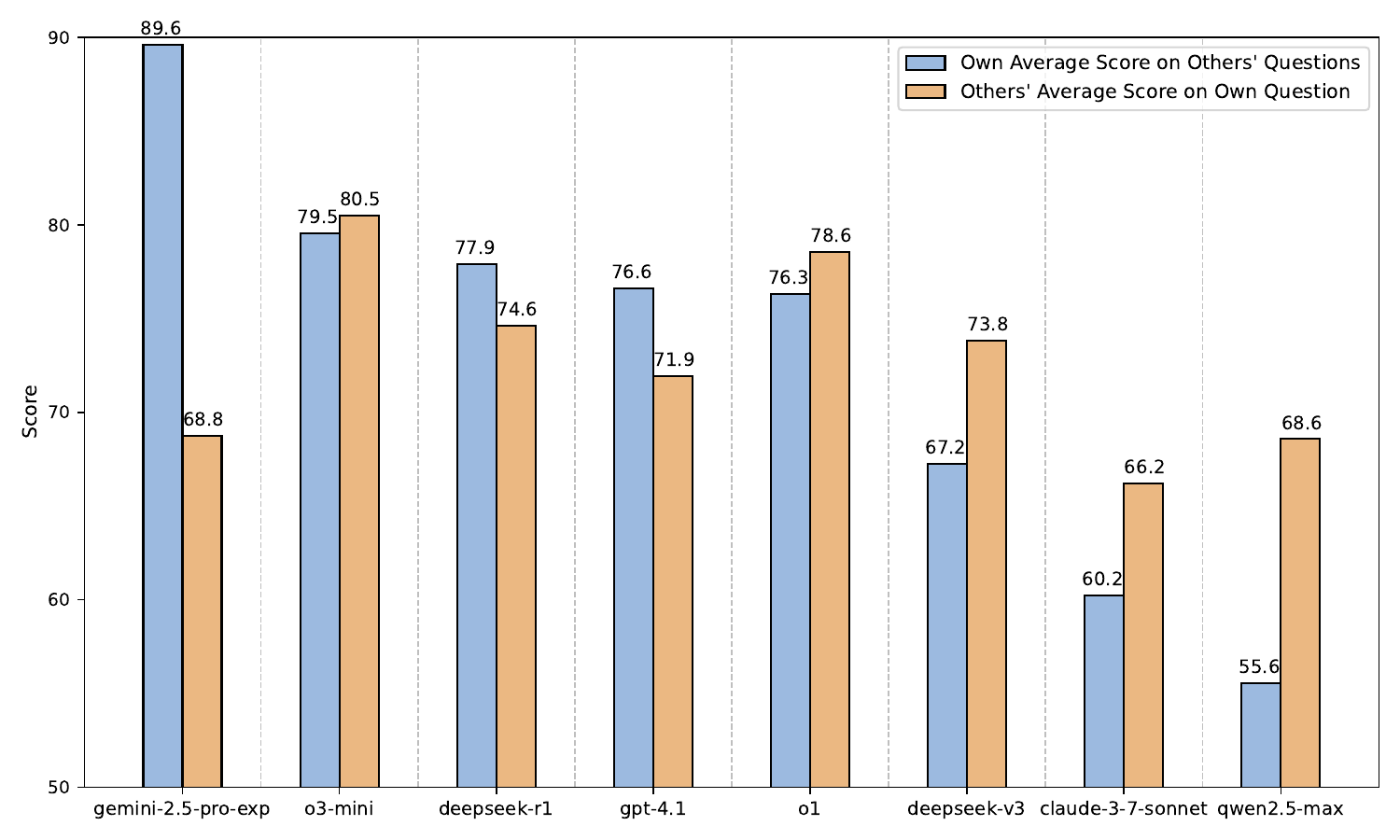} 
\caption{\textbf{Ability comparison of LLMs on programming tasks (question-solving vs. question-setting).} They are measured by Own Average Score on Others' Questions and Others' Average Score on Own Question.}
\label{fig:score}
\end{figure}

As shown in Figure \ref{fig:score}, we conduct a two-dimensional analysis of eight LLMs' overall performance in programming tasks: the average score a LLM receives on other LLMs' questions (Own Average Score on Others' Questions; higher is better)  measures its question-solving ability, while the average score other LLMs receive on that LLM's questions (Others' Average Score on Own Question; lower indicates more difficult questions) reflects its question-setting ability. Statistical details are provided in Appendix \ref{sec:Details for Programming}.

We observe that these two dimensions do not follow a simple linear relationship, but instead exhibit a clear decoupling. This indicates that LLM capability does not follow the common intuition that a stronger solver necessarily sets harder questions. It is similar to the idea that ``a good programmer is not necessarily a good architect'': being skilled at solving questions does not necessarily imply being skilled at designing them. For example, although o3-mini performs well on solving others' questions, the questions it generates are among the easiest for other LLMs to answer, exhibiting a ``strong solver, weak question setter'' profile. In contrast, gemini-2.5-pro-exp maintains high performance on both dimensions, suggesting more balanced capabilities.

This question-setting dimension, automatically induced by the mutual evaluation paradigm, captures asymmetries in LLM cognition that are masked by a single benchmark score. Compared with static benchmarks that report only aggregate scores, this observation reveals deeper cognitive distinctions between LLMs and provides richer empirical evidence for capability comparisons.

\textbf{Finding 5: Higher in-family scores in the \mbox{OpenAI} model family.}

Based on an analysis grouped by developer families, we observe that OpenAI models (gpt-4.1, o3-mini, o1) assign an average score of 79.94 to answers from their own family, but only 70.94 to answers from outside the family ($\Delta = 9$, $p < 0.05$). We refer to this pattern as higher in-family scores in the OpenAI model family. This difference may arise from shared architectures, training data, or output styles, which may lead these models to favor answers from their own family. It may also be partially attributed to the inherently strong reasoning capabilities of the OpenAI models themselves. In contrast, DeepSeek models (deepseek-r1, deepseek-v3) show only a 0.85-point difference between in-family and out-of-family scores ($p > 0.05$), suggesting no notable gap. See Appendix \ref{sec:In-Family} for more details. These results suggest that using a single OpenAI model (e.g., GPT-4) as the evaluator LLM may unintentionally amplify this in-family scoring discrepancy. Meanwhile, LOL can reveal this pattern and mitigate its effect by reducing reliance on any single model.

\begin{figure*}[t]
\centering

\begin{minipage}{0.48\textwidth}
    \centering
    \includegraphics[width=\linewidth]{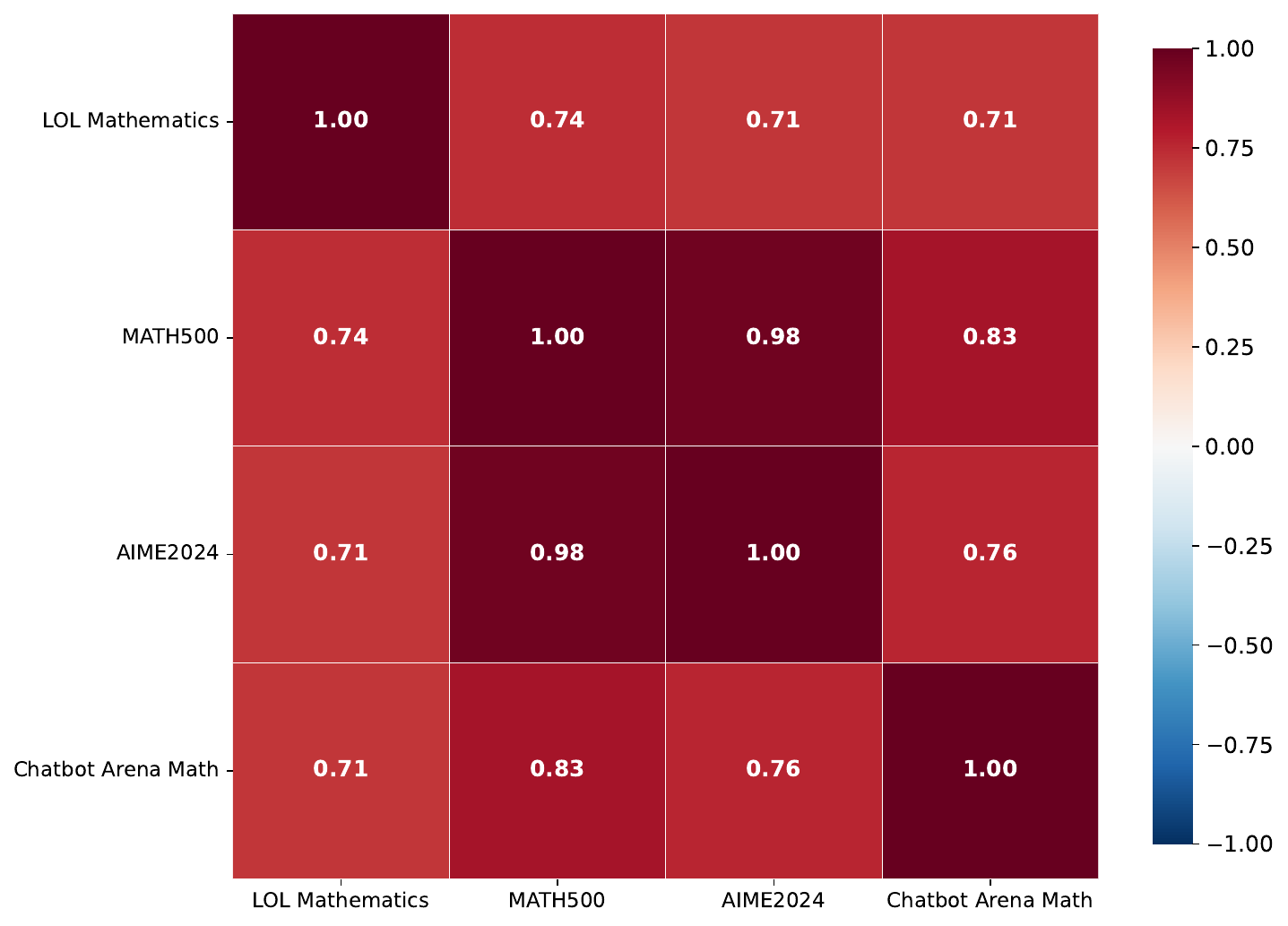}
    \caption*{(a) Mathematics}
\end{minipage}
\hfill
\begin{minipage}{0.48\textwidth}
    \centering
    \includegraphics[width=\linewidth]{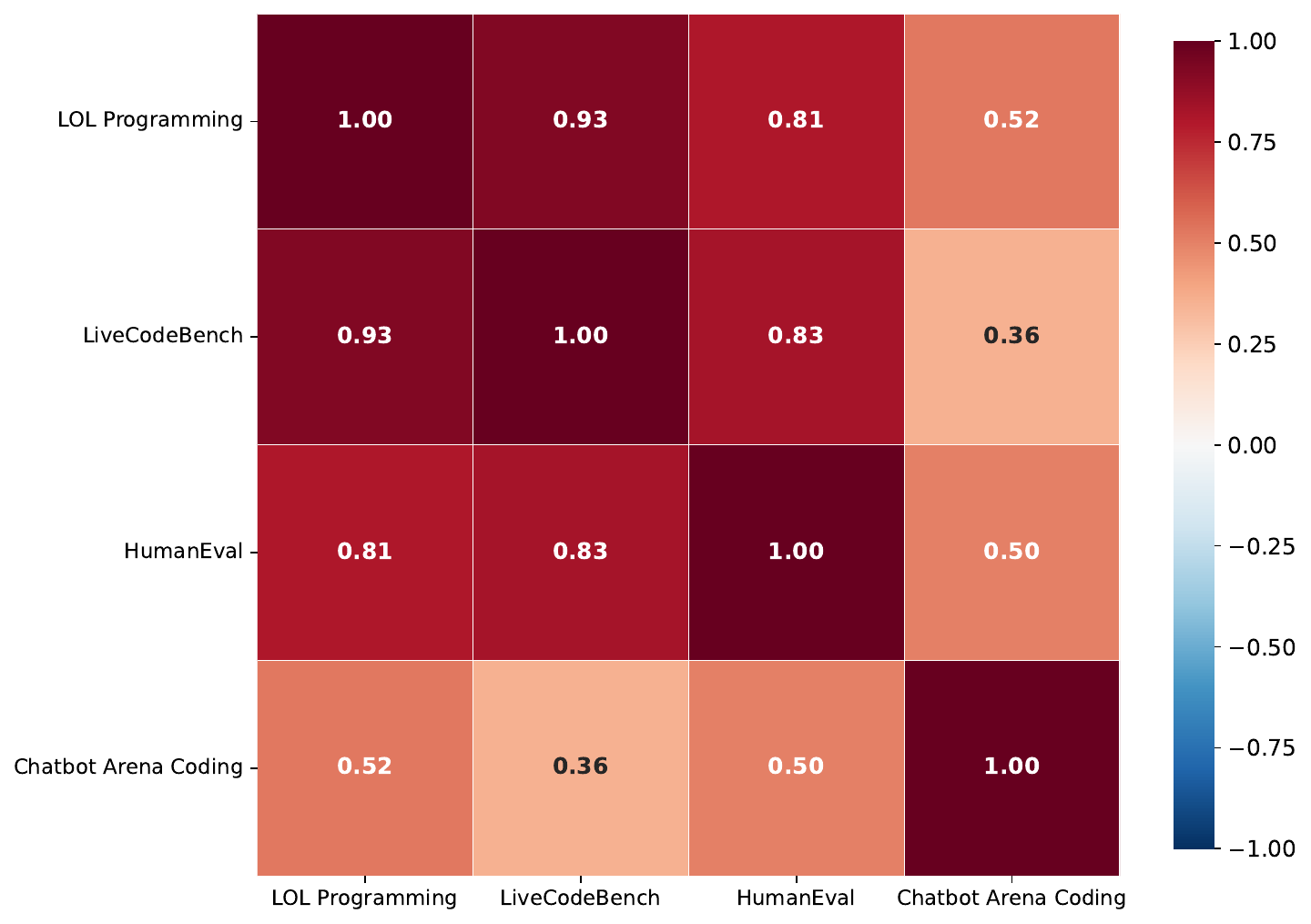}
    \caption*{(b) Programming}
\end{minipage}

\caption{\textbf{Spearman correlation between LOL and Well-Known Benchmarks (Mathematics and Programming).} The heatmaps report Spearman's $\rho$ among LOL and Well-Known Benchmarks (Mathematics: MATH500, AIME2024, Chatbot Arena Math; Programming: LiveCodeBench, HumanEval, Chatbot Arena Coding).}
\label{fig:two_heatmaps}
\end{figure*}

\section{League of LLMs vs. Well-Known Benchmarks}

To verify the external validity of LOL, we compare our mathematics and programming rankings with several well-known benchmarks, including Chatbot Arena, MATH500, AIME2024, LiveCodeBench, and HumanEval. The benchmark rankings are sourced from the official Chatbot Arena website and the Artificial Analysis leaderboard (snapshots taken at the time of retrieval), and we compute Spearman correlation coefficients between LOL and each benchmark ranking.

As shown in Figure \ref{fig:two_heatmaps}, the LOL mathematics ranking shows significant positive correlations with MATH500 ($\rho = 0.74$, $p < 0.05$) and with Chatbot Arena Math and AIME2024 ($\rho = 0.71$, $p < 0.05$). For programming tasks, it likewise remains highly consistent with LiveCodeBench ($\rho = 0.93$, $p < 0.05$) and HumanEval ($\rho = 0.81$, $p < 0.05$). 

These results show that even without relying on human scoring or fixed benchmark datasets, LOL can still produce rankings that are highly consistent with well-known benchmarks, supporting the reliability of LOL's rankings. LOL aligns with the overall trend while still retaining subtle discrepancies, which may stem from differences in evaluation focus and preferences across setups, suggesting that LOL offers a complementary perspective.

\section{Related Work}
The LLM-as-a-Judge line of work \cite{zheng2023judging,pan2024human} is closest to our study. Its core idea is to use a single LLM as a judge to score the answers of other LLMs. However, our work differs fundamentally from this line of work in both goals and design.

LLM-as-a-Judge adopts a single-LLM judge, which may introduce systematic biases, such as self-preference and the higher in-family scores in the OpenAI model family that we empirically observe in Finding 5. In contrast, our study proposes a decentralized closed-loop paradigm based on multi-LLM mutual evaluation with dynamic question generation, which mitigates single-judge bias at the mechanism level. Moreover, while LLM-as-a-Judge evaluates model answers on fixed benchmark datasets, LOL integrates question generation, answering, and mutual evaluation into a single dynamic and self-consistent closed loop. As a result, LOL reduces reliance on fixed benchmark datasets by autonomously generating and updating its own test instances.

\section{Discussion}
\textbf{Q1: Are the LLM-generated questions and reference answers reliable?}

We require questioner LLMs to provide reference answers simultaneously, ensuring each question has a clear, verifiable answer. All questions and reference answers were validated independently by at least three human annotators without modification, and we did not identify any unsolvable or logically inconsistent items.

Moreover, LOL does not rely solely on a single reference answer: even if a reference answer is imperfect, multiple LLMs can independently evaluate answers, reducing reliance on any single reference. Meanwhile, with multiple LLMs rotating as questioners, even low-discrimination questions can be diluted over rounds through mutual evaluation. To further address concerns about potential answer dependency, we conducted an ablation by removing the reference answer from the evaluator prompt. Although absolute scores exhibit some fluctuation, the final model rankings remain largely consistent, preserving all main conclusions of the paper.

Empirically, robustness is supported by high Top-$k$ consistency and statistical tests, as well as strong correlations with well-known benchmarks. Together, these results suggest that the resulting rankings are reliable.

\textbf{Q2: If the evaluation LLMs may have biases or errors, how can the results be ensured to be objective and reasonable?} 

We prohibit self-evaluation and adopt a ``mutual evaluation with reference answer'' mechanism to reduce bias. To further reduce idiosyncratic errors from any single evaluator, we aggregate judgments from multiple LLMs across multiple rounds. Even if misjudgments occur, errors can be mitigated through mutual evaluation. 

We have performed five independent runs for each experiment, and the resulting rankings are highly consistent across runs, indicating that the evaluation is robust and reproducible. LLMs can exhibit more professional domain knowledge and objective evaluation standards than most crowd users. Notably, we position LOL as a valuable complement rather than a replacement for existing evaluation paradigms. While LLM evaluators may still be imperfect, LOL provides a scalable, benchmark-free paradigm for more objective and reproducible evaluation.

\textbf{Q3: What is the operational cost of LOL, and how can its scalability be maintained under high computational complexity?}

Table~\ref{tab:cost} reports the average token consumption for each model in a full run of the main experiment. The average cost per complete experiment (averaged across runs) is approximately \$1.87 for mathematics tasks and \$3.29 for programming tasks.

\begin{table*}[!htbp]
\centering
\small
\setlength{\tabcolsep}{4pt}
\begin{tabular}{lcccc}
\hline
Model & Mathematics Input & Mathematics Output & Programming Input & Programming Output\\
\hline
gpt-4.1 & 27444 & 11501 & 94345 & 14107\\
o3-mini & 29128 & 43270 & 94016 & 62587\\
o1 & 29501 & 61358 & 95051 & 94492\\
claude-3-7-sonnet & 32273 & 15223 & 103860 & 27166\\
deepseek-r1 & 30568 & 146624 & 84495 & 300803\\
deepseek-v3 & 33422 & 18425 & 128434 & 11965\\
qwen2.5-max & 28853 & 7260 & 96431 & 15559\\
gemini-2.5-pro-exp & 26379 & 150988 & 96148 & 207968\\
\hline
\end{tabular}
\caption{Average token consumption per model across runs for mathematics and programming tasks.}
\label{tab:cost}
\end{table*}

Given the computational cost of full mutual evaluation, we conduct a sparse Elo replay experiment to explore the feasibility of reducing complexity. Specifically, we randomly sample 1\%--50\% of comparisons from the full pairwise comparison pool and independently repeat the process 100 times to reconstruct the rankings. The results show that, even with only 30\% of the comparisons, the recovered rankings remain highly consistent with those from full evaluation (mathematics: Spearman's $\rho=1.00$; programming: Spearman's $\rho=0.93$). This demonstrates that LOL can substantially reduce computational cost via sparse sampling while preserving high ranking reliability, thereby exhibiting strong scalability.

\textbf{Q4: How is the originality of the generated questions verified?}

We conduct an originality analysis by measuring both lexical and semantic similarity between the generated questions and widely used benchmarks, using 5-gram overlap for lexical similarity and sentence embeddings from all-MiniLM-L6-v2 \cite{wang2020minilm} for semantic similarity. For mathematics, comparisons are conducted against a total of 9,322 questions from MATH500, AIME2024, and GSM8K, yielding an average 5-gram overlap of 5.11\% and a semantic similarity of 0.4856; for programming (1,219 questions), comparisons against LiveCodeBench and HumanEval yield an average 5-gram overlap of 5.90\% and a semantic similarity of 0.5730. These results indicate low lexical and semantic similarity to existing benchmarks, suggesting that the generated questions are not simple paraphrases of prior datasets.

\textbf{Q5: How can repeated or highly similar questions be avoided in long-term operation?}

To mitigate repetition, we introduce two mechanisms. A League Memory maintains an embedding-based memory bank of historical questions, where newly generated questions are compared against their Top-$k$ nearest neighbors and those exceeding a similarity threshold are filtered and regenerated. In addition, the system dynamically schedules topics through a seasonal rotation mechanism to diversify topic distribution and mitigate local repetition.

\section{Conclusion}
In this paper, we propose League of LLMs (LOL), a novel benchmark-free evaluation paradigm that organizes multiple LLMs into a self-governed league for multi-round mutual evaluation. LOL integrates four core criteria to mitigate key limitations of mainstream evaluation paradigms, including data contamination, opaque operation, and subjective preferences. Experiments on eight mainstream LLMs in mathematics and programming show that LOL can effectively differentiate LLM capabilities and produce highly consistent internal ranking across runs (Top-$k$ consistency $= 70.7\%$). Beyond ranking, LOL surfaces empirical findings that are difficult for traditional paradigms to capture in practice, including ``memorization-based answering'' and higher in-family scores in the OpenAI model family. We publicly release our framework and code to support the evaluation of newly released LLMs and to serve as a valuable complement to the current LLM evaluation ecosystem.

\section*{Limitations}

\textbf{Tasks and Models.} Due to space limitations, we validate LOL only in mathematics and programming. Future work will extend LOL to more tasks (e.g., reasoning, dialogue, and open-ended writing) and more models.

\textbf{Evaluator reliability and bias.} LOL reduces reliance on a single evaluator model through multi-LLM mutual evaluation and prohibits self-evaluation. However, correlated errors may still arise when multiple models share similar preferences or blind spots. The reference answer mechanism may be less effective for open-ended tasks. In addition, although LOL does not require human scoring or adjudication, we still performed lightweight human validation of the generated questions and reference answers in our experiments as a sanity check. Therefore, future work may incorporate stronger automated verification and calibration, complemented by lightweight expert spot-checking, to further mitigate biases.

\textbf{Scalability and cost.} LOL supports continual extension. When adding a new model, mutual-evaluation outcomes can be converted into pairwise comparison signals and integrated via an online rating scheme such as Elo, allowing the model to enter the leaderboard through a limited number of ``matches'' against a subset of existing models rather than re-running the full mutual evaluation with all existing models (conceptually similar to Chatbot Arena's pairwise-comparison aggregation). As the model pool grows, maintaining efficient computation and question diversity becomes increasingly important. Future work can explore more efficient scheduling and sampling strategies (e.g., selective evaluation or adaptive rounds) to reduce cost while maintaining stability.

\section*{Acknowledgments}

We sincerely thank the anonymous reviewers and the Area Chair for their constructive feedback and valuable suggestions. In particular, we are grateful for the insightful comments that led us to conduct additional experiments, including the ablation study without reference answers and the sparse Elo experiment, etc. These suggestions have significantly strengthened the paper.


\bibliography{custom}

@article{liang2023holistic,
title={Holistic Evaluation of Language Models},
author={Percy Liang and Rishi Bommasani and Tony Lee and Dimitris Tsipras and Dilara Soylu and Michihiro Yasunaga and Yian Zhang and Deepak Narayanan and Yuhuai Wu and Ananya Kumar and Benjamin Newman and Binhang Yuan and Bobby Yan and Ce Zhang and Christian Cosgrove and Christopher D Manning and Christopher Re and Diana Acosta-Navas and Drew A. Hudson and Eric Zelikman and Esin Durmus and Faisal Ladhak and Frieda Rong and Hongyu Ren and Huaxiu Yao and Jue WANG and Keshav Santhanam and Laurel Orr and Lucia Zheng and Mert Yuksekgonul and Mirac Suzgun and Nathan Kim and Neel Guha and Niladri S. Chatterji and Omar Khattab and Peter Henderson and Qian Huang and Ryan Andrew Chi and Sang Michael Xie and Shibani Santurkar and Surya Ganguli and Tatsunori Hashimoto and Thomas Icard and Tianyi Zhang and Vishrav Chaudhary and William Wang and Xuechen Li and Yifan Mai and Yuhui Zhang and Yuta Koreeda},
journal={Transactions on Machine Learning Research},
issn={2835-8856},
year={2023},
url={https://openreview.net/forum?id=iO4LZibEqW},
note={Featured Certification, Expert Certification, Outstanding Certification}
}

@article{hendrycks2021measuring,
  title={Measuring mathematical problem solving with the math dataset},
  author={Hendrycks, Dan and Burns, Collin and Kadavath, Saurav and Arora, Akul and Basart, Steven and Tang, Eric and Song, Dawn and Steinhardt, Jacob},
  journal={arXiv preprint arXiv:2103.03874},
  year={2021}
}

@article{jain2024livecodebench,
  title={Livecodebench: Holistic and contamination free evaluation of large language models for code},
  author={Jain, Naman and Han, King and Gu, Alex and Li, Wen-Ding and Yan, Fanjia and Zhang, Tianjun and Wang, Sida and Solar-Lezama, Armando and Sen, Koushik and Stoica, Ion},
  journal={arXiv preprint arXiv:2403.07974},
  year={2024}
}

@article{hendrycks2020measuring,
  title={Measuring massive multitask language understanding},
  author={Hendrycks, Dan and Burns, Collin and Basart, Steven and Zou, Andy and Mazeika, Mantas and Song, Dawn and Steinhardt, Jacob},
  journal={arXiv preprint arXiv:2009.03300},
  year={2020}
}

@article{mirzadeh2024gsm,
  title={Gsm-symbolic: Understanding the limitations of mathematical reasoning in large language models},
  author={Mirzadeh, Iman and Alizadeh, Keivan and Shahrokhi, Hooman and Tuzel, Oncel and Bengio, Samy and Farajtabar, Mehrdad},
  journal={arXiv preprint arXiv:2410.05229},
  year={2024}
}

@article{xie2024large,
  title={Do Large Language Models Truly Grasp Mathematics? An Empirical Exploration From Cognitive Psychology},
  author={Xie, Wei and Ma, Shuoyoucheng and Wang, Zhenhua and Wang, Enze and Chen, Kai and Sun, Xiaobing and Wang, Baosheng},
  journal={arXiv preprint arXiv:2410.14979},
  year={2024}
}

@inproceedings{rein2024gpqa,
  title={Gpqa: A graduate-level google-proof q\&a benchmark},
  author={Rein, David and Hou, Betty Li and Stickland, Asa Cooper and Petty, Jackson and Pang, Richard Yuanzhe and Dirani, Julien and Michael, Julian and Bowman, Samuel R},
  booktitle={First Conference on Language Modeling},
  year={2024}
}

@article{wang2019superglue,
  title={Superglue: A stickier benchmark for general-purpose language understanding systems},
  author={Wang, Alex and Pruksachatkun, Yada and Nangia, Nikita and Singh, Amanpreet and Michael, Julian and Hill, Felix and Levy, Omer and Bowman, Samuel},
  journal={Advances in neural information processing systems},
  volume={32},
  year={2019}
}

@inproceedings{casper2024black,
  title={Black-box access is insufficient for rigorous ai audits},
  author={Casper, Stephen and Ezell, Carson and Siegmann, Charlotte and Kolt, Noam and Curtis, Taylor Lynn and Bucknall, Benjamin and Haupt, Andreas and Wei, Kevin and Scheurer, J{\'e}r{\'e}my and Hobbhahn, Marius and others},
  booktitle={Proceedings of the 2024 ACM Conference on Fairness, Accountability, and Transparency},
  pages={2254--2272},
  year={2024}
}

@inproceedings{chiang2024chatbot,
  title={Chatbot arena: An open platform for evaluating llms by human preference},
  author={Chiang, Wei-Lin and Zheng, Lianmin and Sheng, Ying and Angelopoulos, Anastasios Nikolas and Li, Tianle and Li, Dacheng and Zhu, Banghua and Zhang, Hao and Jordan, Michael and Gonzalez, Joseph E and others},
  booktitle={Forty-first International Conference on Machine Learning},
  year={2024}
}

@misc{sentimentarena2025,
    title = {Introducing Sentiment Control: Disentagling Sentiment and Substance},
    url = {https://blog.lmarena.ai/blog/2025/sentiment-control/},
    urldate = {2025-12-15},
    author = {Chen, Connor and Chiang, Wei-Lin and Li, Tianle and Angelopoulos, Anastasios N.},
    month = {April},
    year = {2025}
}

@article{li2024does,
  title={Does style matter? disentangling style and substance in chatbot arena},
  author={Li, Tianle and Angelopoulos, Anastasios and Chiang, Wei-Lin},
  journal={LMSYS Blog},
  year={2024}
}

@article{achiam2023gpt,
  title={Gpt-4 technical report},
  author={Achiam, Josh and Adler, Steven and Agarwal, Sandhini and Ahmad, Lama and Akkaya, Ilge and Aleman, Florencia Leoni and Almeida, Diogo and Altenschmidt, Janko and Altman, Sam and Anadkat, Shyamal and others},
  journal={arXiv preprint arXiv:2303.08774},
  year={2023}
}

@article{team2023gemini,
  title={Gemini: a family of highly capable multimodal models},
  author={Team, Gemini and Anil, Rohan and Borgeaud, Sebastian and Alayrac, Jean-Baptiste and Yu, Jiahui and Soricut, Radu and Schalkwyk, Johan and Dai, Andrew M and Hauth, Anja and Millican, Katie and others},
  journal={arXiv preprint arXiv:2312.11805},
  year={2023}
}

@article{yang2023rethinking,
  title={Rethinking benchmark and contamination for language models with rephrased samples},
  author={Yang, Shuo and Chiang, Wei-Lin and Zheng, Lianmin and Gonzalez, Joseph E and Stoica, Ion},
  journal={arXiv preprint arXiv:2311.04850},
  year={2023}
}

@article{chen2024humans,
  title={Humans or llms as the judge? a study on judgement biases},
  author={Chen, Guiming Hardy and Chen, Shunian and Liu, Ziche and Jiang, Feng and Wang, Benyou},
  journal={arXiv preprint arXiv:2402.10669},
  year={2024}
}

@article{cobbe2021training,
  title={Training verifiers to solve math word problems},
  author={Cobbe, Karl and Kosaraju, Vineet and Bavarian, Mohammad and Chen, Mark and Jun, Heewoo and Kaiser, Lukasz and Plappert, Matthias and Tworek, Jerry and Hilton, Jacob and Nakano, Reiichiro and others},
  journal={arXiv preprint arXiv:2110.14168},
  year={2021}
}

@article{chen2021evaluating,
  title={Evaluating large language models trained on code},
  author={Chen, Mark and Tworek, Jerry and Jun, Heewoo and Yuan, Qiming and Pinto, Henrique Ponde De Oliveira and Kaplan, Jared and Edwards, Harri and Burda, Yuri and Joseph, Nicholas and Brockman, Greg and others},
  journal={arXiv preprint arXiv:2107.03374},
  year={2021}
}

@inproceedings{xu2022systematic,
  title={A systematic evaluation of large language models of code},
  author={Xu, Frank F and Alon, Uri and Neubig, Graham and Hellendoorn, Vincent Josua},
  booktitle={Proceedings of the 6th ACM SIGPLAN international symposium on machine programming},
  pages={1--10},
  year={2022}
}

@article{nejjar2025llms,
  title={Llms for science: Usage for code generation and data analysis},
  author={Nejjar, Mohamed and Zacharias, Luca and Stiehle, Fabian and Weber, Ingo},
  journal={Journal of Software: Evolution and Process},
  volume={37},
  number={1},
  pages={e2723},
  year={2025},
  publisher={Wiley Online Library}
}

@article{kasneci2023chatgpt,
  title={ChatGPT for good? On opportunities and challenges of large language models for education},
  author={Kasneci, Enkelejda and Se{\ss}ler, Kathrin and K{\"u}chemann, Stefan and Bannert, Maria and Dementieva, Daryna and Fischer, Frank and Gasser, Urs and Groh, Georg and G{\"u}nnemann, Stephan and H{\"u}llermeier, Eyke and others},
  journal={Learning and individual differences},
  volume={103},
  pages={102274},
  year={2023},
  publisher={Elsevier}
}

@article{chang2024survey,
  title={A survey on evaluation of large language models},
  author={Chang, Yupeng and Wang, Xu and Wang, Jindong and Wu, Yuan and Yang, Linyi and Zhu, Kaijie and Chen, Hao and Yi, Xiaoyuan and Wang, Cunxiang and Wang, Yidong and others},
  journal={ACM transactions on intelligent systems and technology},
  volume={15},
  number={3},
  pages={1--45},
  year={2024},
  publisher={ACM New York, NY}
}

@article{boyko2023interdisciplinary,
  title={An interdisciplinary outlook on large language models for scientific research},
  author={Boyko, James and Cohen, Joseph and Fox, Nathan and Veiga, Maria Han and Li, Jennifer I and Liu, Jing and Modenesi, Bernardo and Rauch, Andreas H and Reid, Kenneth N and Tribedi, Soumi and others},
  journal={arXiv preprint arXiv:2311.04929},
  year={2023}
}

@article{ahn2024large,
  title={Large language models for mathematical reasoning: Progresses and challenges},
  author={Ahn, Janice and Verma, Rishu and Lou, Renze and Liu, Di and Zhang, Rui and Yin, Wenpeng},
  journal={arXiv preprint arXiv:2402.00157},
  year={2024}
}

@article{huang2023competition,
  title={Competition-level problems are effective llm evaluators},
  author={Huang, Yiming and Lin, Zhenghao and Liu, Xiao and Gong, Yeyun and Lu, Shuai and Lei, Fangyu and Liang, Yaobo and Shen, Yelong and Lin, Chen and Duan, Nan and others},
  journal={arXiv preprint arXiv:2312.02143},
  year={2023}
}

@article{cheng2025survey,
  title={A survey on data contamination for large language models},
  author={Cheng, Yuxing and Chang, Yi and Wu, Yuan},
  journal={arXiv preprint arXiv:2502.14425},
  year={2025}
}

@article{xu2024benchmark,
  title={Benchmark data contamination of large language models: A survey},
  author={Xu, Cheng and Guan, Shuhao and Greene, Derek and Kechadi, M and others},
  journal={arXiv preprint arXiv:2406.04244},
  year={2024}
}

@article{deng2023investigating,
  title={Investigating data contamination in modern benchmarks for large language models},
  author={Deng, Chunyuan and Zhao, Yilun and Tang, Xiangru and Gerstein, Mark and Cohan, Arman},
  journal={arXiv preprint arXiv:2311.09783},
  year={2023}
}

@article{sainz2023nlp,
  title={NLP evaluation in trouble: On the need to measure LLM data contamination for each benchmark},
  author={Sainz, Oscar and Campos, Jon Ander and Garc{\'\i}a-Ferrero, Iker and Etxaniz, Julen and de Lacalle, Oier Lopez and Agirre, Eneko},
  journal={arXiv preprint arXiv:2310.18018},
  year={2023}
}

@article{azerbayev2023llemma,
  title={Llemma: An open language model for mathematics},
  author={Azerbayev, Zhangir and Schoelkopf, Hailey and Paster, Keiran and Santos, Marco Dos and McAleer, Stephen and Jiang, Albert Q and Deng, Jia and Biderman, Stella and Welleck, Sean},
  journal={arXiv preprint arXiv:2310.10631},
  year={2023}
}

@inproceedings{lu2023survey,
  title={A Survey of Deep Learning for Mathematical Reasoning},
  author={Lu, Pan and Qiu, Liang and Yu, Wenhao and Welleck, Sean and Chang, Kai-Wei},
  booktitle={ACL (1)},
  year={2023}
}

@article{luo2023wizardmath,
  title={Wizardmath: Empowering mathematical reasoning for large language models via reinforced evol-instruct},
  author={Luo, Haipeng and Sun, Qingfeng and Xu, Can and Zhao, Pu and Lou, Jianguang and Tao, Chongyang and Geng, Xiubo and Lin, Qingwei and Chen, Shifeng and Zhang, Dongmei},
  journal={arXiv preprint arXiv:2308.09583},
  year={2023}
}

@article{yu2023metamath,
  title={Metamath: Bootstrap your own mathematical questions for large language models},
  author={Yu, Longhui and Jiang, Weisen and Shi, Han and Yu, Jincheng and Liu, Zhengying and Zhang, Yu and Kwok, James T and Li, Zhenguo and Weller, Adrian and Liu, Weiyang},
  journal={arXiv preprint arXiv:2309.12284},
  year={2023}
}

@article{ge2023openagi,
  title={Openagi: When llm meets domain experts},
  author={Ge, Yingqiang and Hua, Wenyue and Mei, Kai and Tan, Juntao and Xu, Shuyuan and Li, Zelong and Zhang, Yongfeng and others},
  journal={Advances in Neural Information Processing Systems},
  volume={36},
  pages={5539--5568},
  year={2023}
}

@article{touvron2023llama,
  title={Llama: Open and efficient foundation language models},
  author={Touvron, Hugo and Lavril, Thibaut and Izacard, Gautier and Martinet, Xavier and Lachaux, Marie-Anne and Lacroix, Timoth{\'e}e and Rozi{\`e}re, Baptiste and Goyal, Naman and Hambro, Eric and Azhar, Faisal and others},
  journal={arXiv preprint arXiv:2302.13971},
  year={2023}
}

@article{vaswani2017attention,
  title={Attention is all you need},
  author={Vaswani, Ashish and Shazeer, Noam and Parmar, Niki and Uszkoreit, Jakob and Jones, Llion and Gomez, Aidan N and Kaiser, {\L}ukasz and Polosukhin, Illia},
  journal={Advances in neural information processing systems},
  volume={30},
  year={2017}
}

@article{li2024gsm,
  title={Gsm-plus: A comprehensive benchmark for evaluating the robustness of llms as mathematical problem solvers},
  author={Li, Qintong and Cui, Leyang and Zhao, Xueliang and Kong, Lingpeng and Bi, Wei},
  journal={arXiv preprint arXiv:2402.19255},
  year={2024}
}

@article{liu2023llms,
  title={LLMs as narcissistic evaluators: When ego inflates evaluation scores},
  author={Liu, Yiqi and Moosavi, Nafise Sadat and Lin, Chenghua},
  journal={arXiv preprint arXiv:2311.09766},
  year={2023}
}

@article{ye2024justice,
  title={Justice or prejudice? quantifying biases in llm-as-a-judge},
  author={Ye, Jiayi and Wang, Yanbo and Huang, Yue and Chen, Dongping and Zhang, Qihui and Moniz, Nuno and Gao, Tian and Geyer, Werner and Huang, Chao and Chen, Pin-Yu and others},
  journal={arXiv preprint arXiv:2410.02736},
  year={2024}
}

@article{borda1781memoire,
  title={M{\'e}moire sur les {\'e}lections au scrutin: Histoire de l’Acad{\'e}mie Royale des Sciences},
  author={Borda, J-C de},
  journal={Paris, France},
  volume={12},
  number={2},
  year={1781}
}

@article{zheng2023judging,
  title={Judging llm-as-a-judge with mt-bench and chatbot arena},
  author={Zheng, Lianmin and Chiang, Wei-Lin and Sheng, Ying and Zhuang, Siyuan and Wu, Zhanghao and Zhuang, Yonghao and Lin, Zi and Li, Zhuohan and Li, Dacheng and Xing, Eric and others},
  journal={Advances in neural information processing systems},
  volume={36},
  pages={46595--46623},
  year={2023}
}

@article{pan2024human,
  title={Human-Centered Design Recommendations for LLM-as-a-judge},
  author={Pan, Qian and Ashktorab, Zahra and Desmond, Michael and Cooper, Martin Santillan and Johnson, James and Nair, Rahul and Daly, Elizabeth and Geyer, Werner},
  journal={arXiv preprint arXiv:2407.03479},
  year={2024}
}

@inproceedings{moore2024assessing,
  title={Assessing Educational Quality: Comparative Analysis of Crowdsourced, Expert, and AI-Driven Rubric Applications},
  author={Moore, Steven and Bier, Norman and Stamper, John},
  booktitle={Proceedings of the AAAI Conference on Human Computation and Crowdsourcing},
  volume={12},
  pages={115--125},
  year={2024}
}

@inproceedings{zhang2025lists,
  title={From lists to emojis: How format bias affects model alignment},
  author={Zhang, Xuanchang and Xiong, Wei and Chen, Lichang and Zhou, Tianyi and Huang, Heng and Zhang, Tong},
  booktitle={Proceedings of the 63rd Annual Meeting of the Association for Computational Linguistics (Volume 1: Long Papers)},
  pages={26940--26961},
  year={2025}
}

@misc{huiyan_api,
  title   = {Huiyan {API}},
  author  = {{Huiyan}},
  year    = {2023},
  url     = {https://api.huiyan-ai.cn/},
  urldate = {2025-12-15}
}

@article{wang2020minilm,
  title={Minilm: Deep self-attention distillation for task-agnostic compression of pre-trained transformers},
  author={Wang, Wenhui and Wei, Furu and Dong, Li and Bao, Hangbo and Yang, Nan and Zhou, Ming},
  journal={Advances in neural information processing systems},
  volume={33},
  pages={5776--5788},
  year={2020}
}

\appendix
\section{Statistical Details for Mathematics}
\label{sec:Details for Mathematics}

Table \ref{tab:math_overall_borda} summarizes the Borda scores received by each model in Mathematics, including the mean and 95\% confidence interval (CI) of the mean.

\begin{table}[!htbp]
\centering
\small
\setlength{\tabcolsep}{4pt}
\begin{tabular}{lcc}
\hline
Model & Mean & 95\% CI \\
\hline
gemini-2.5-pro-exp   & 5.1388 & [4.9509, 5.3102] \\
deepseek-r1          & 4.1918 & [3.9959, 4.3796] \\
gpt-4.1              & 3.8776 & [3.7020, 4.0409] \\
o3-mini              & 3.5714 & [3.4000, 3.7347] \\
o1                   & 3.4204 & [3.2408, 3.6000] \\
deepseek-v3          & 2.8531 & [2.6735, 3.0491] \\
claude-3-7-sonnet    & 2.5510 & [2.3673, 2.7593] \\
qwen2.5-max          & 1.8204 & [1.6531, 1.9918] \\
\hline
\end{tabular}
\caption{\textbf{Mathematics overall summary (Borda score in [0, 6]).} Mean Borda score received by each model, with 95\% CI of the mean.}
\label{tab:math_overall_borda}
\end{table}

Table \ref{tab:math_tukey_ns} reports only the four non-significant ($p>0.05$) pairs from Tukey HSD with adjusted $p$-values (p-adj); all remaining pairs are significant ($p<0.05$).

\begin{table}[!htbp]
\centering
\small
\setlength{\tabcolsep}{4pt}
\begin{tabular}{llc}
\hline
Model 1 & Model 2 & p-adj \\
\hline
claude-3-7-sonnet & deepseek-v3 & 0.3162 \\
deepseek-r1       & gpt-4.1     & 0.2659 \\
gpt-4.1           & o3-mini     & 0.2989 \\
o1                & o3-mini     & 0.9502 \\
\hline
\end{tabular}
\caption{\textbf{Non-significant Tukey HSD pairs in Mathematics.} We report only the non-significant pairs ($p>0.05$); all remaining pairs are significant ($p<0.05$).}
\label{tab:math_tukey_ns}
\end{table}

\section{Statistical Details for Programming}
\label{sec:Details for Programming}

Table \ref{tab:prog_overall_score} summarizes the overall scores received by each model in programming, reporting the mean and 95\% confidence interval (CI) of the mean.

\begin{table}[!htbp]
\centering
\small
\setlength{\tabcolsep}{4pt}
\begin{tabular}{lcc}
\hline
Model & Mean & 95\% CI \\
\hline
gemini-2.5-pro-exp   & 89.6000 & [88.0933, 91.0205] \\
o3-mini              & 79.5469 & [77.0893, 82.0612] \\
deepseek-r1          & 77.9061 & [75.0488, 80.3430] \\
gpt-4.1              & 76.6082 & [73.6727, 79.1968] \\
o1                   & 76.3347 & [73.5750, 79.0083] \\
deepseek-v3          & 67.2490 & [64.0241, 70.1633] \\
claude-3-7-sonnet    & 60.2163 & [57.6361, 62.7634] \\
qwen2.5-max           & 55.5510 & [52.4143, 58.6740] \\
\hline
\end{tabular}
\caption{\textbf{Programming overall summary (absolute score in [0, 100]).} Mean score received by each model, with 95\% CI of the mean.}
\label{tab:prog_overall_score}
\end{table}

\section{Score Heatmaps}
\label{sec:Score Heatmaps}
\begin{figure*}[!p]
\centering
\includegraphics[width=0.98\textwidth]{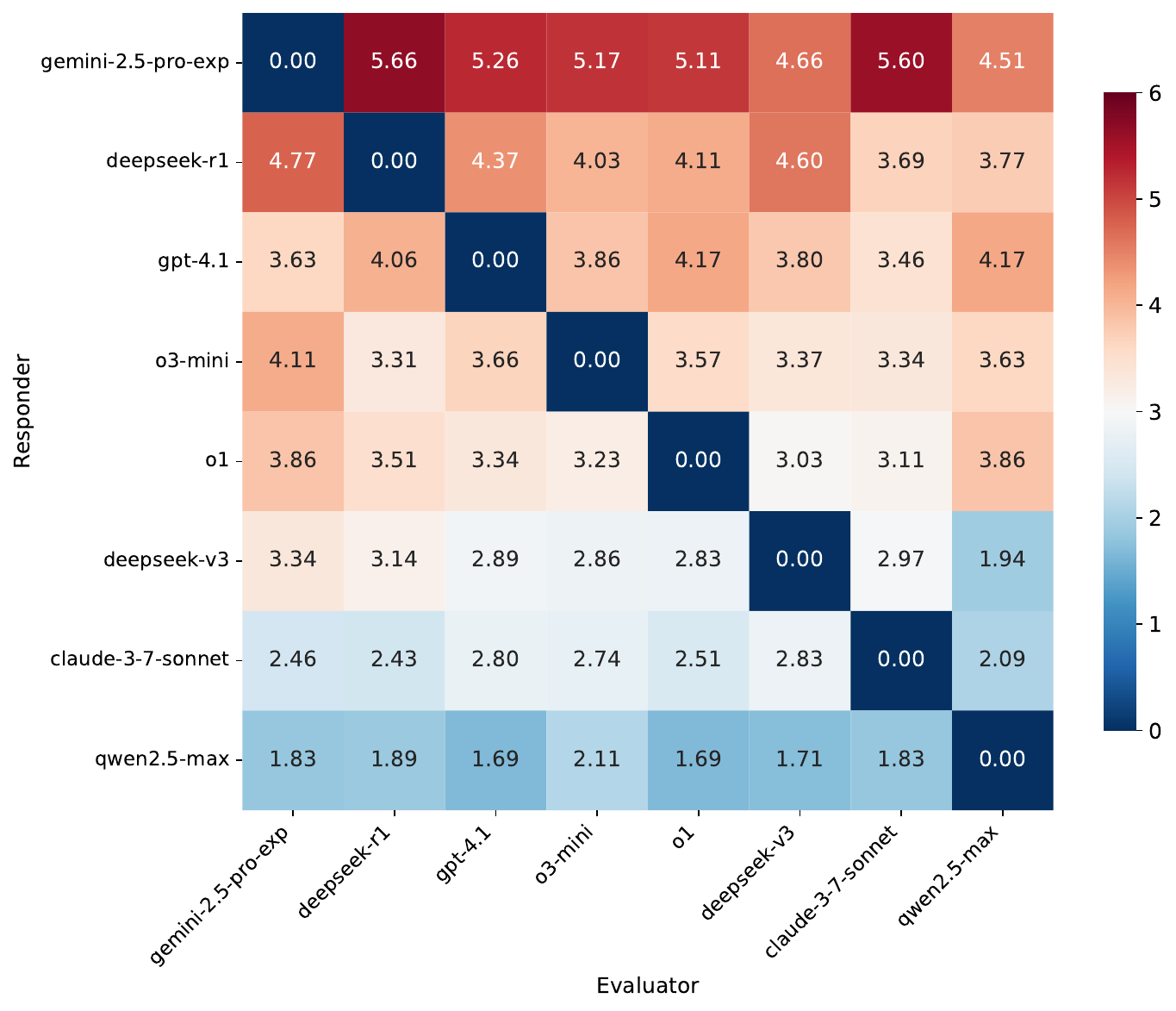} 
\caption{\textbf{Mathematics Scores.} Rows are responders and columns are evaluators; each cell shows the mean score.}
\label{fig:score heatmaps 1}
\end{figure*}

\begin{figure*}[!p]
\centering
\includegraphics[width=0.98\textwidth]{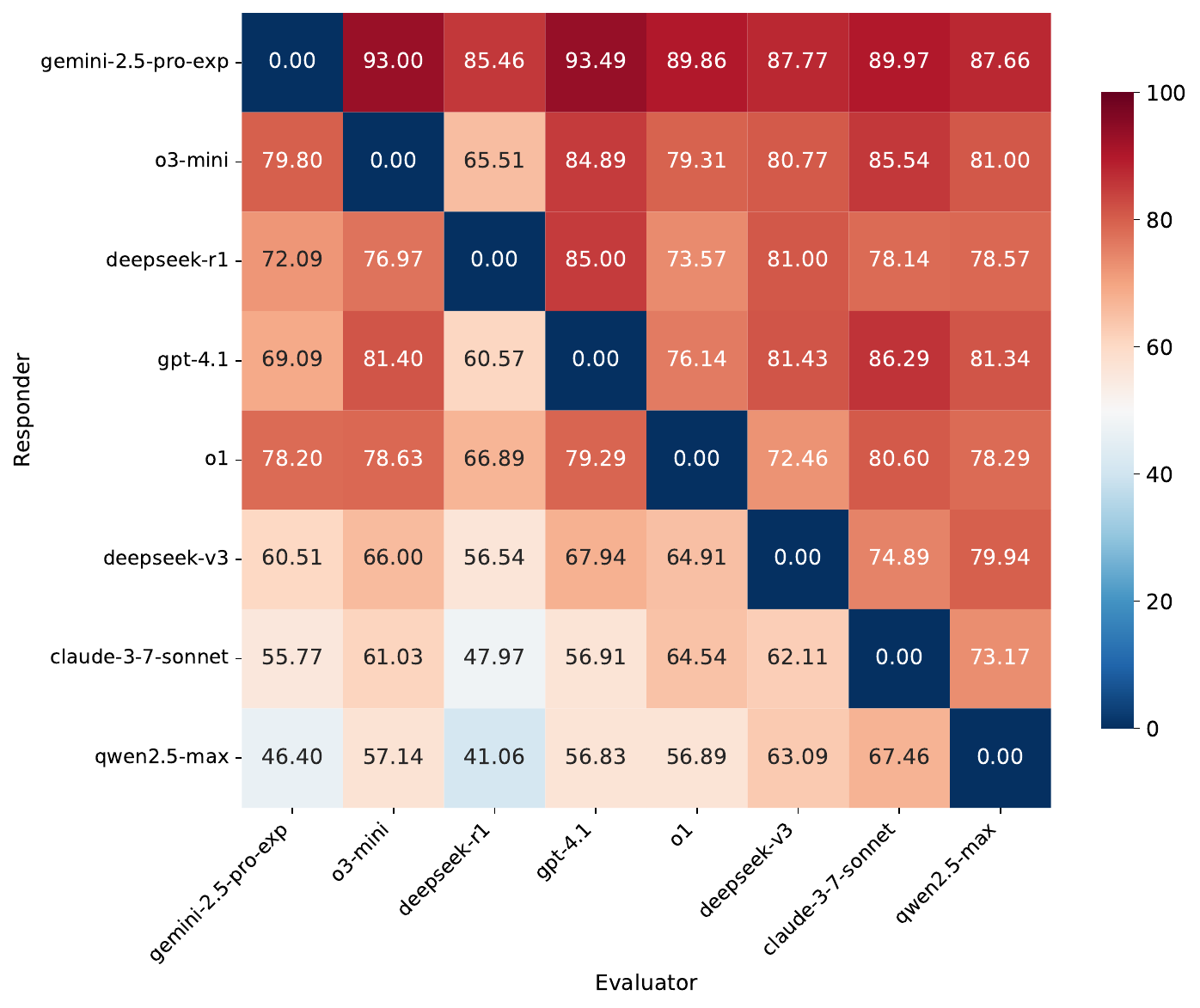} 
\caption{\textbf{Programming Scores.} Rows are responders and columns are evaluators; each cell shows the mean score.}
\label{fig:score heatmaps 2}
\end{figure*}

We refer to the answering LLM as the responder here. Figures \ref{fig:score heatmaps 1} and \ref{fig:score heatmaps 2} summarize the mean scores for each evaluator-responder pair. The diagonal is set to 0 since self-evaluation is prohibited.

\begin{table*}[!t]
\centering
\small
\begin{tabular}{lccccc}
\hline
Family & Models & In-family & Out-of-family & $\Delta$ & $p$-value\\
\hline
OpenAI   & gpt-4.1, o3-mini, o1 & 79.9429 & 70.9390 & 9.0038 & 0.026770 \\
DeepSeek & deepseek-r1, deepseek-v3 & 68.7714 & 67.9238 & 0.8476 & 0.956624 \\
\hline
\end{tabular}
\caption{\textbf{Developer-family in-family scoring differences (in-family vs. out-of-family).} We report the mean scores, the difference $\Delta$, and Welch's two-sided $t$-test $p$-values.}
\label{tab:In-Family table}
\end{table*}

\FloatBarrier

\section{Top-k Consistency Details}
\label{sec:Top-k}

Table \ref{tab:topk_consistency_math} summarizes Top-$k$ consistency in Mathematics across five independent runs, reporting the mean overlap and the 95\% confidence interval (CI) for each $k$.

\begin{table}[!htbp]
\centering
\small
\setlength{\tabcolsep}{4pt}
\begin{tabular}{ccc}
\hline
$k$ & Top-$k$ value & 95\% CI \\
\hline
1 & 0.4330 & [0.3857, 0.4705] \\
2 & 0.5469 & [0.5147, 0.5746] \\
3 & 0.6339 & [0.6190, 0.6485] \\
4 & 0.7067 & [0.6998, 0.7125] \\
5 & 0.7762 & [0.7698, 0.7834] \\
6 & 0.8396 & [0.8393, 0.8402] \\
7 & 1.0000 & [1.0000, 1.0000] \\
\hline
\end{tabular}
\caption{\textbf{Mathematics Top-$k$ consistency.} Mean overlap of Top-$k$ model sets across five independent runs, with 95\% CI of the mean.}
\label{tab:topk_consistency_math}
\end{table}

Table \ref{tab:std} reports the standard deviation of each model's overall mathematics score across five independent runs. 

\begin{table}[!h]
\centering
\small
\setlength{\tabcolsep}{4pt}
\begin{tabular}{lcc}
\hline
Model & Std (Borda) & Std (100-point) \\
\hline
gemini-2.5-pro-exp            & 0.3932 & 6.5532 \\
deepseek-r1                   & 0.3635 & 6.0588 \\
gpt-4.1                       & 0.2625 & 4.3757 \\
o3-mini                       & 0.3319 & 5.5318 \\
o1                            & 0.1966 & 3.2766 \\
deepseek-v3                   & 0.3770 & 6.2829 \\
claude-3-7-sonnet             & 0.6454 & 10.7560 \\
qwen2.5-max                   & 0.4055 & 6.7575 \\
\hline
\end{tabular}

\caption{\textbf{Mathematics score dispersion across runs.} Standard deviation (Std) of each model's overall score across five independent runs. We report Std on the Borda scale ([0, 6]) and after linear rescaling to a 100-point scale (Std$_{100}$ = Std$_{\text{Borda}} \times 100/6$).}
\label{tab:std}
\end{table}

\section{Higher In-Family Scores in Developer Families}
\label{sec:In-Family}

Table \ref{tab:In-Family table} compares evaluators' scores on in-family vs. out-of-family answering LLMs.

\textbf{Difference.} $\Delta =$ (in-family mean scores) $-$ (out-of-family mean scores).

\textbf{Sample unit.} We pool all eligible evaluation scores into two sets (in-family vs. out-of-family) for each developer family and compute their means.

\textbf{Statistical test.} We use Welch's two-sided $t$-test (unequal variances) to compare in-family and out-of-family scores for each developer family.

\section{Case Study}
\label{sec:Case Study}

\subsection{Case of Finding 1}
\label{sec:Case of Finding 1}
\begin{figure*}[!htbp]
\centering
\includegraphics[width=0.98\textwidth]{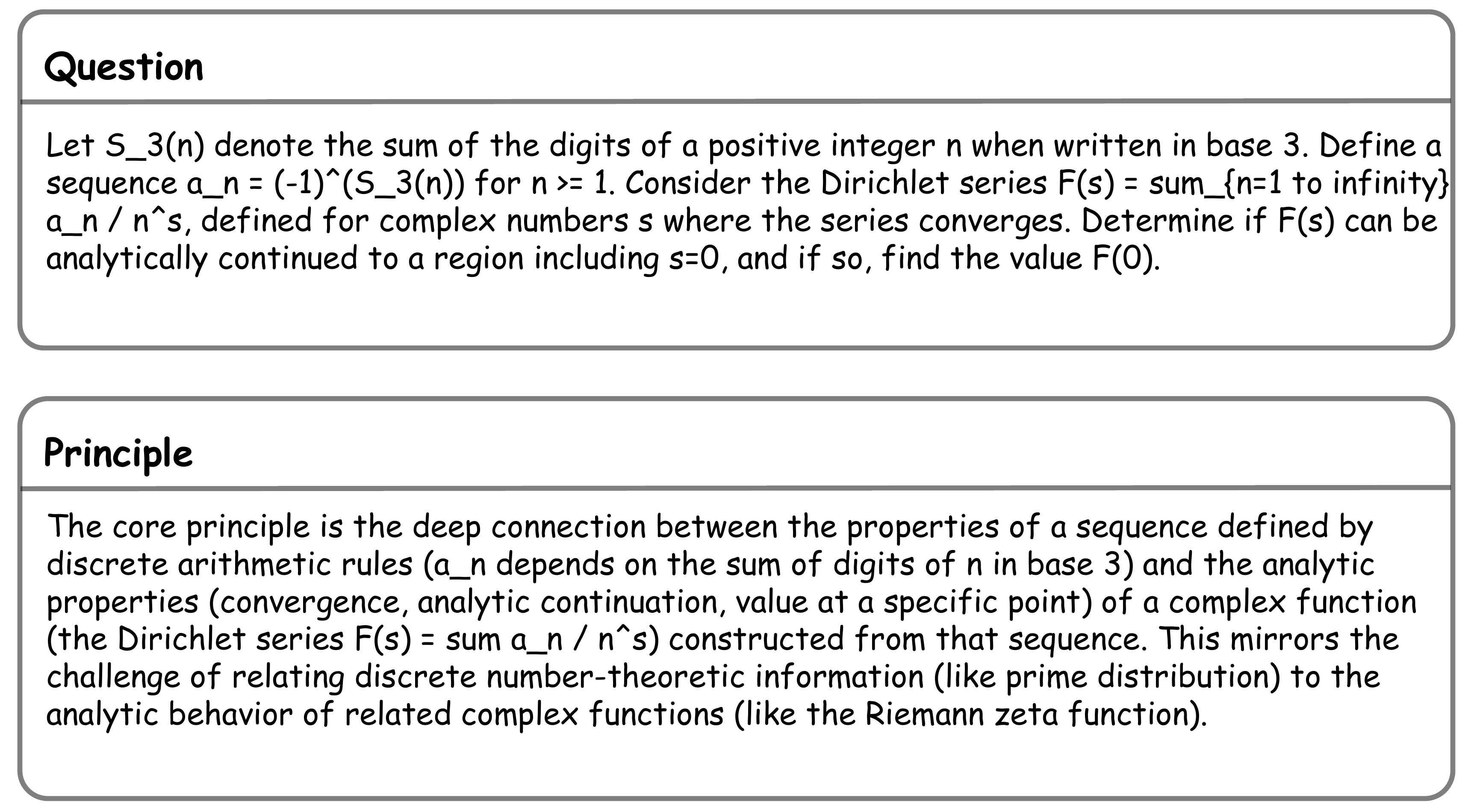}
\caption{\textbf{Case study for Finding 1.} Question and principle.}
\label{fig:Finding11}
\end{figure*}

\begin{figure*}[!htbp]
\centering
\includegraphics[width=0.98\textwidth]{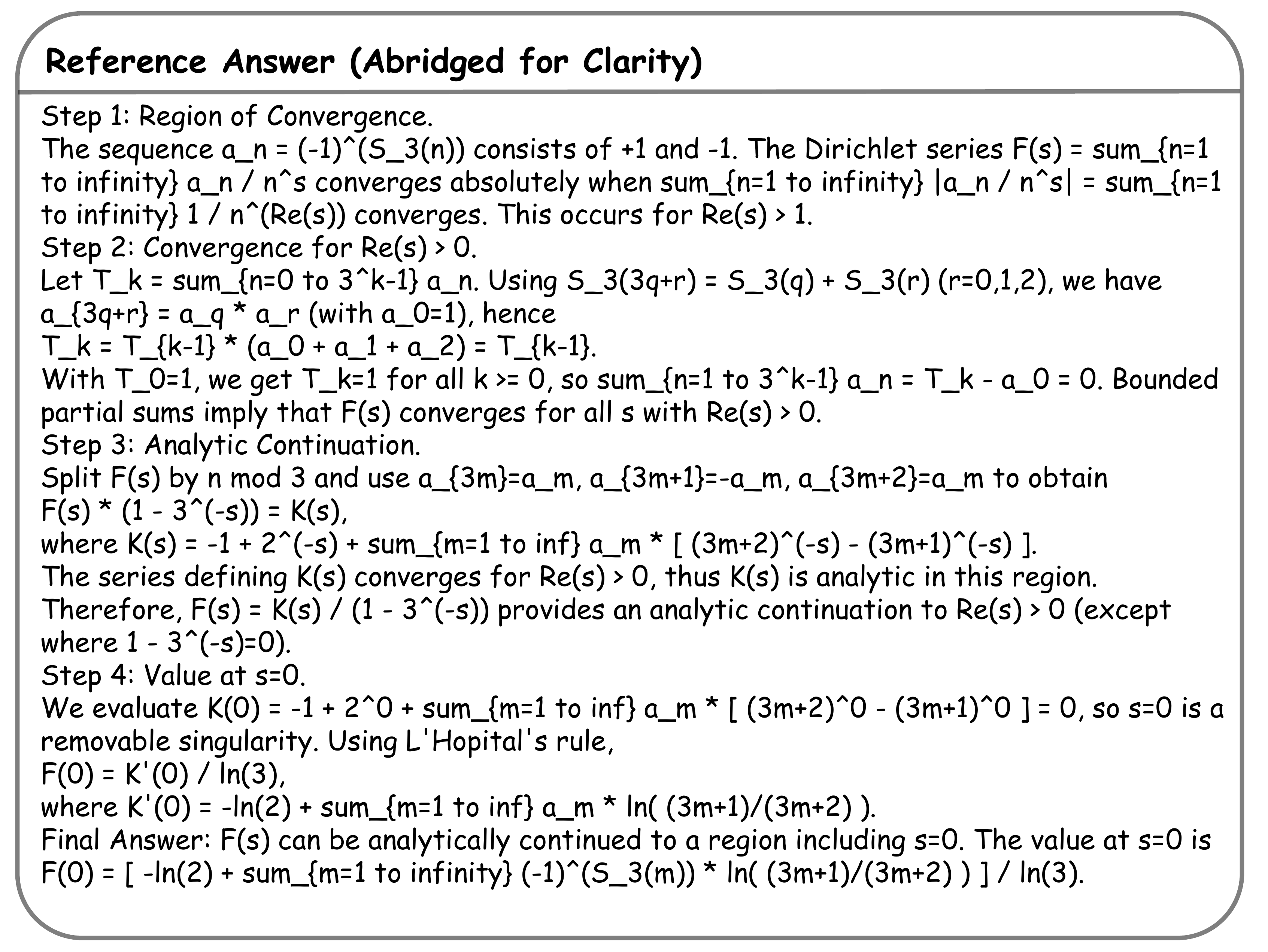}
\caption{\textbf{Case study for Finding 1.} Reference answer (abridged for clarity).}
\label{fig:Finding12}
\end{figure*}

\noindent See Figures \ref{fig:Finding11} and \ref{fig:Finding12}. An example from gemini-2.5-pro-exp-03-25 illustrates professional-level mathematical question generation.

\subsection{Case of Finding 2}
\label{sec:Case of Finding 2}
\begin{figure*}[!htbp]
\centering
\includegraphics[width=0.98\textwidth]{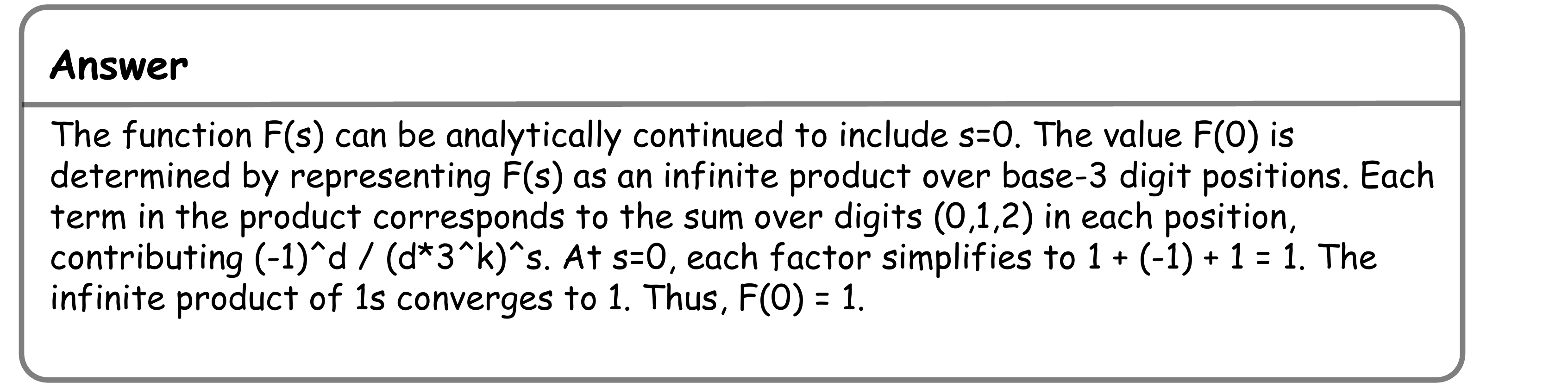}
\caption{\textbf{Case study for Finding 2.} Example answer exhibiting ``memorization-based answering''.}
\label{fig:Finding2}
\end{figure*}

Deepseek-r1 exhibits ``memorization-based answering'' behavior that mimics reasoning. Figure \ref{fig:Finding2} presents a representative example.

\section{Prompts}
\label{sec:Full Prompts}
\subsection{Prompts for Mathematics}
\label{sec:Math Prompts}
\begin{figure*}[!htbp]
\centering
\includegraphics[width=0.98\textwidth]{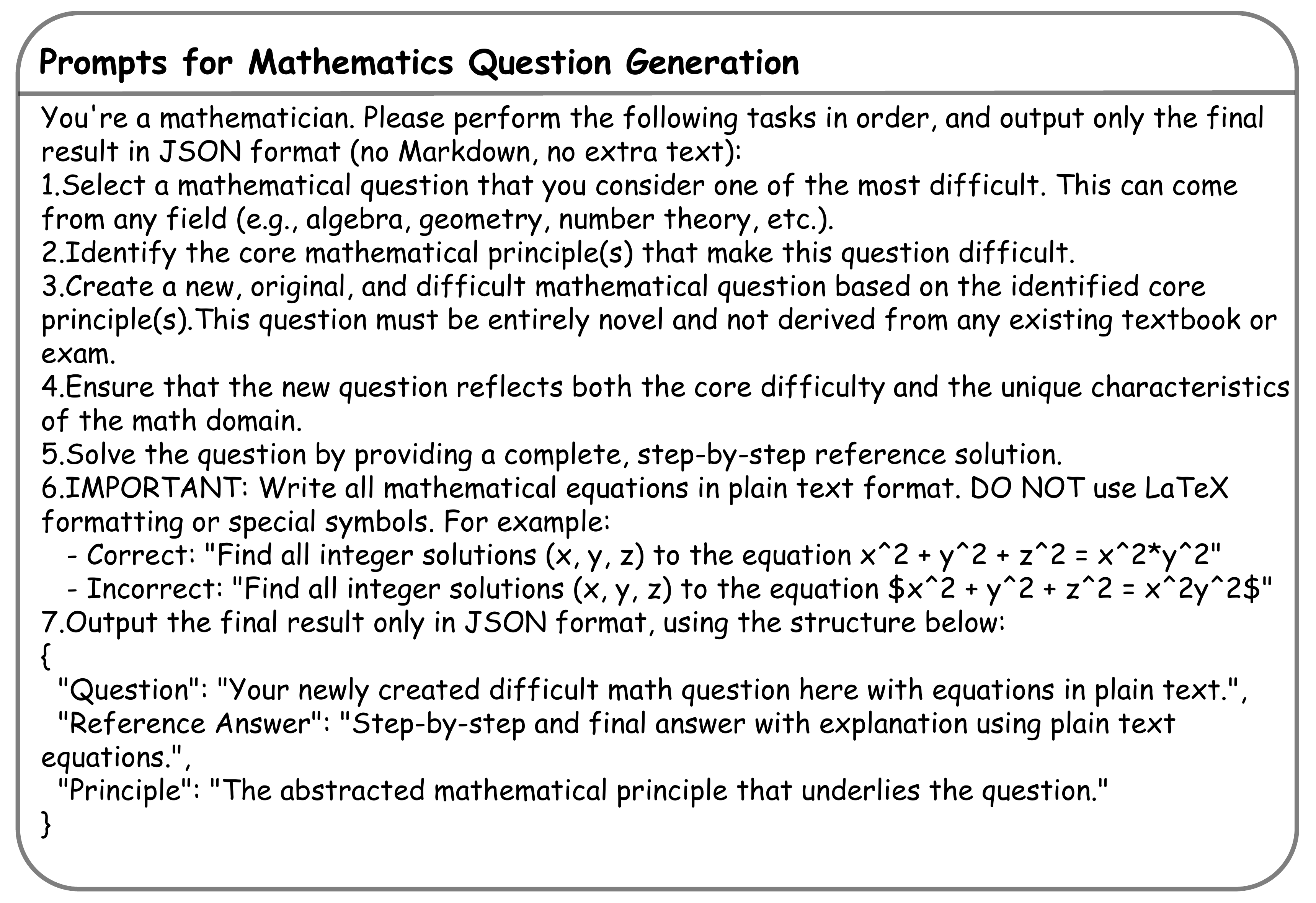}
\caption{Prompts for Mathematics Question Generation.}
\label{fig:math_prompt_qgen}
\end{figure*}

\begin{figure*}[!htbp]
\centering
\includegraphics[width=0.98\textwidth]{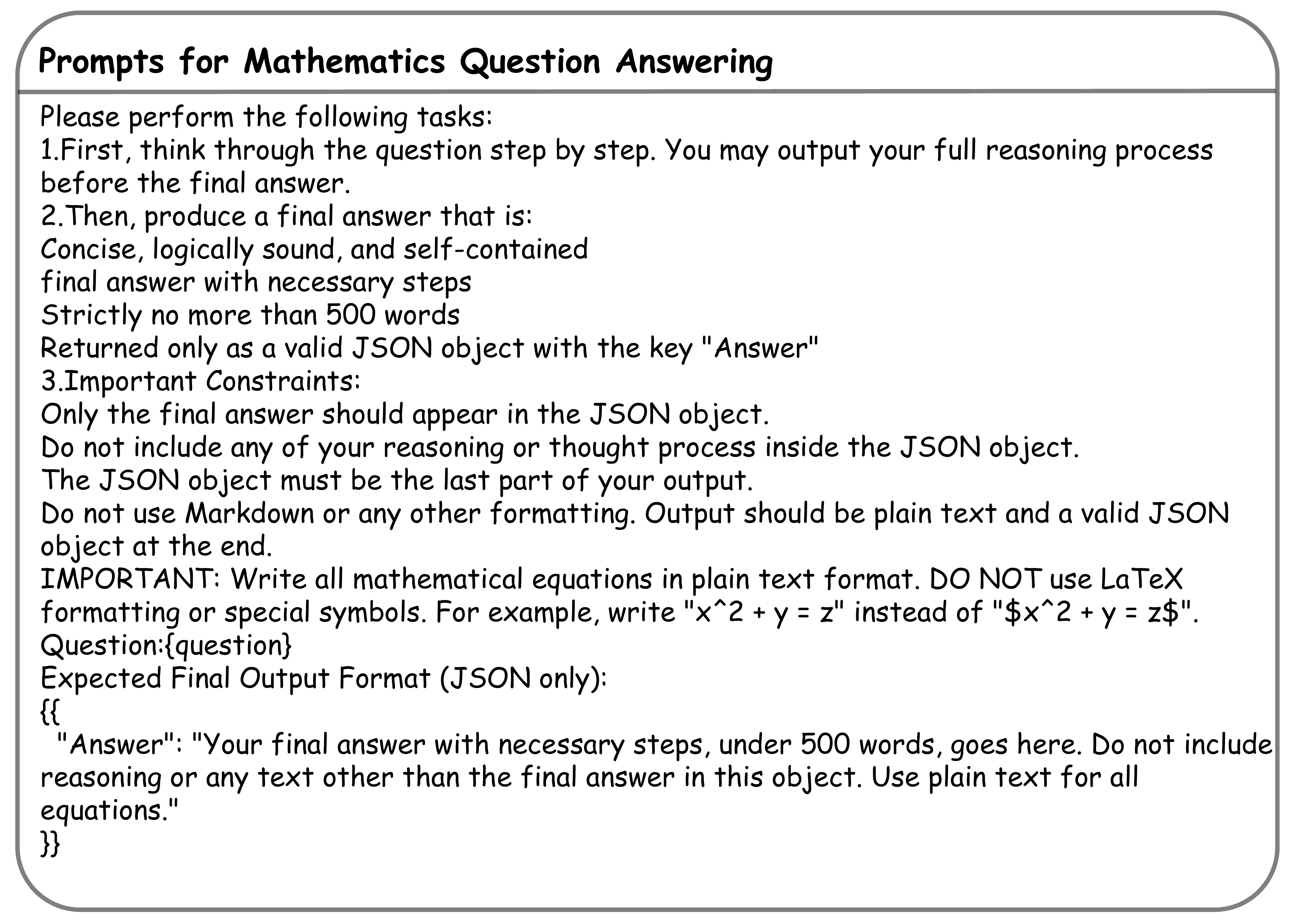}
\caption{Prompts for Mathematics Question Answering.}
\label{fig:math_prompt_ans}
\end{figure*}

\begin{figure*}[!htbp]
\centering
\includegraphics[width=0.98\textwidth]{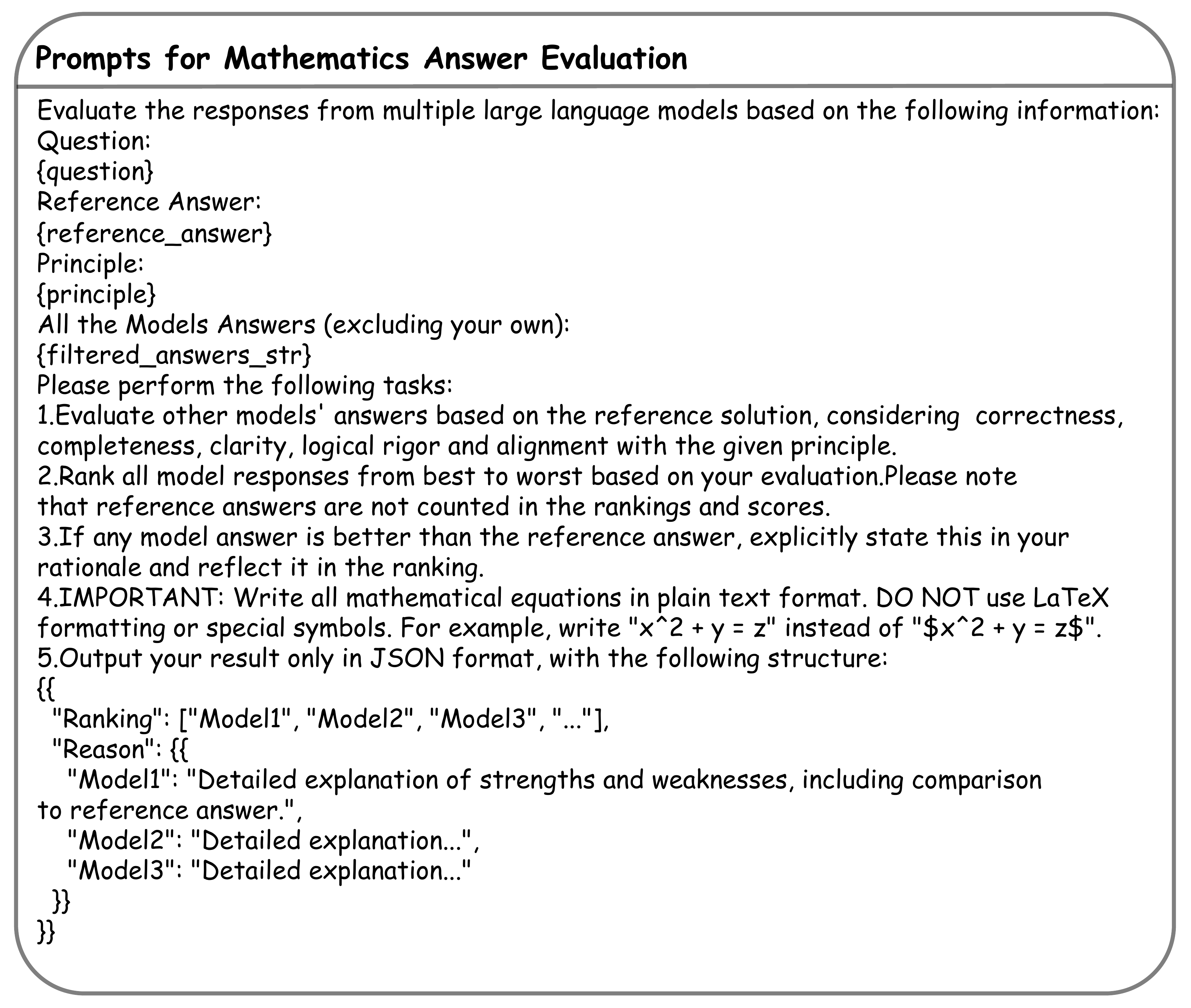}
\caption{Prompts for Mathematics Answer Evaluation.}
\label{fig:math_prompt_eval}
\end{figure*}

The full prompts used for Mathematics include question generation, question answering, and answer evaluation.
The corresponding prompts are presented in Figures \ref{fig:math_prompt_qgen}--\ref{fig:math_prompt_eval}.

\subsection{Prompts for Programming}
\label{sec:Programming Prompts}
\begin{figure*}[!htbp]
\centering
\includegraphics[width=0.98\textwidth]{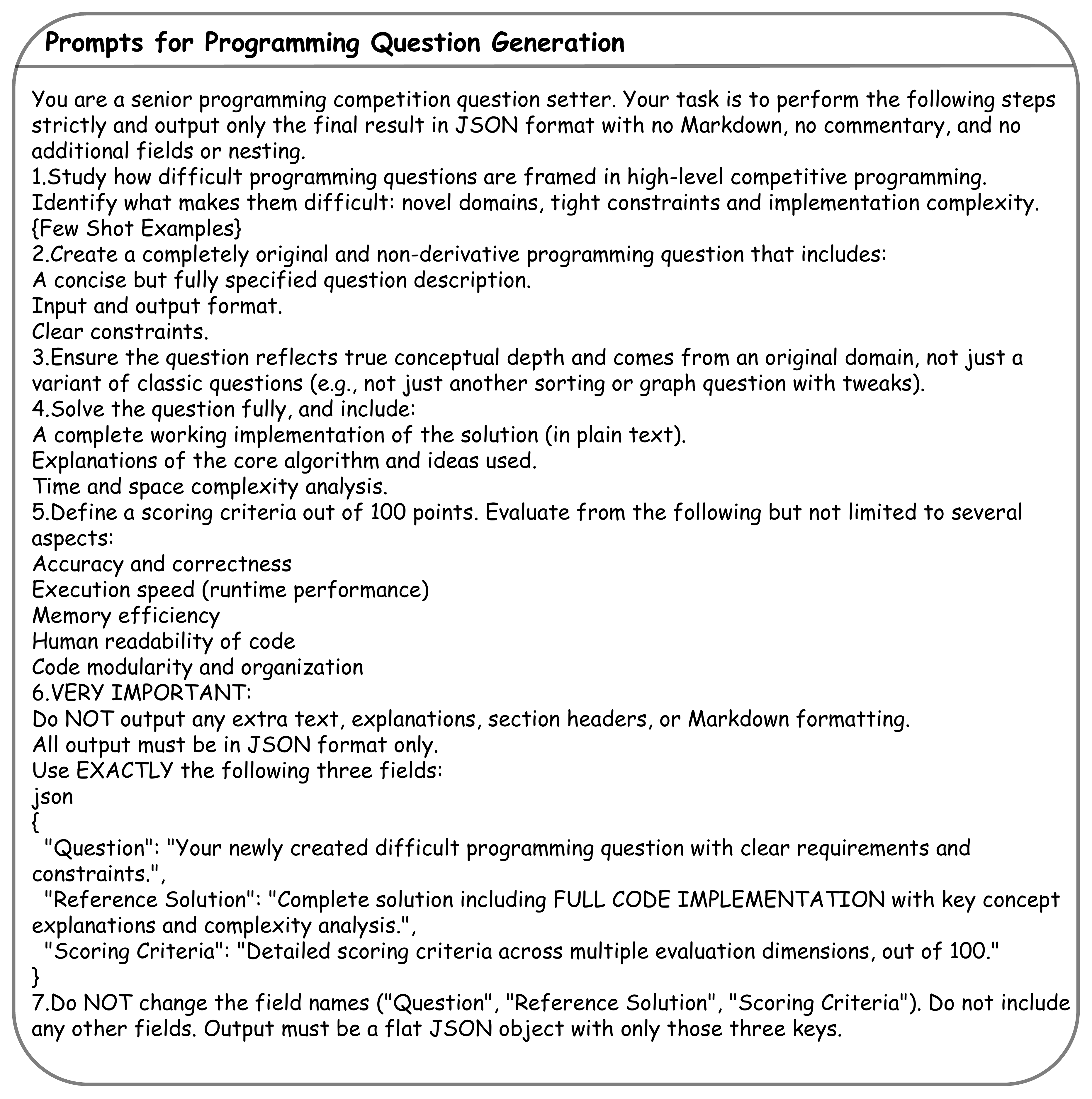}
\caption{Prompts for Programming Question Generation.}
\label{fig:prog_prompt_qgen}
\end{figure*}

\begin{figure*}[!htbp]
\centering
\includegraphics[width=0.98\textwidth]{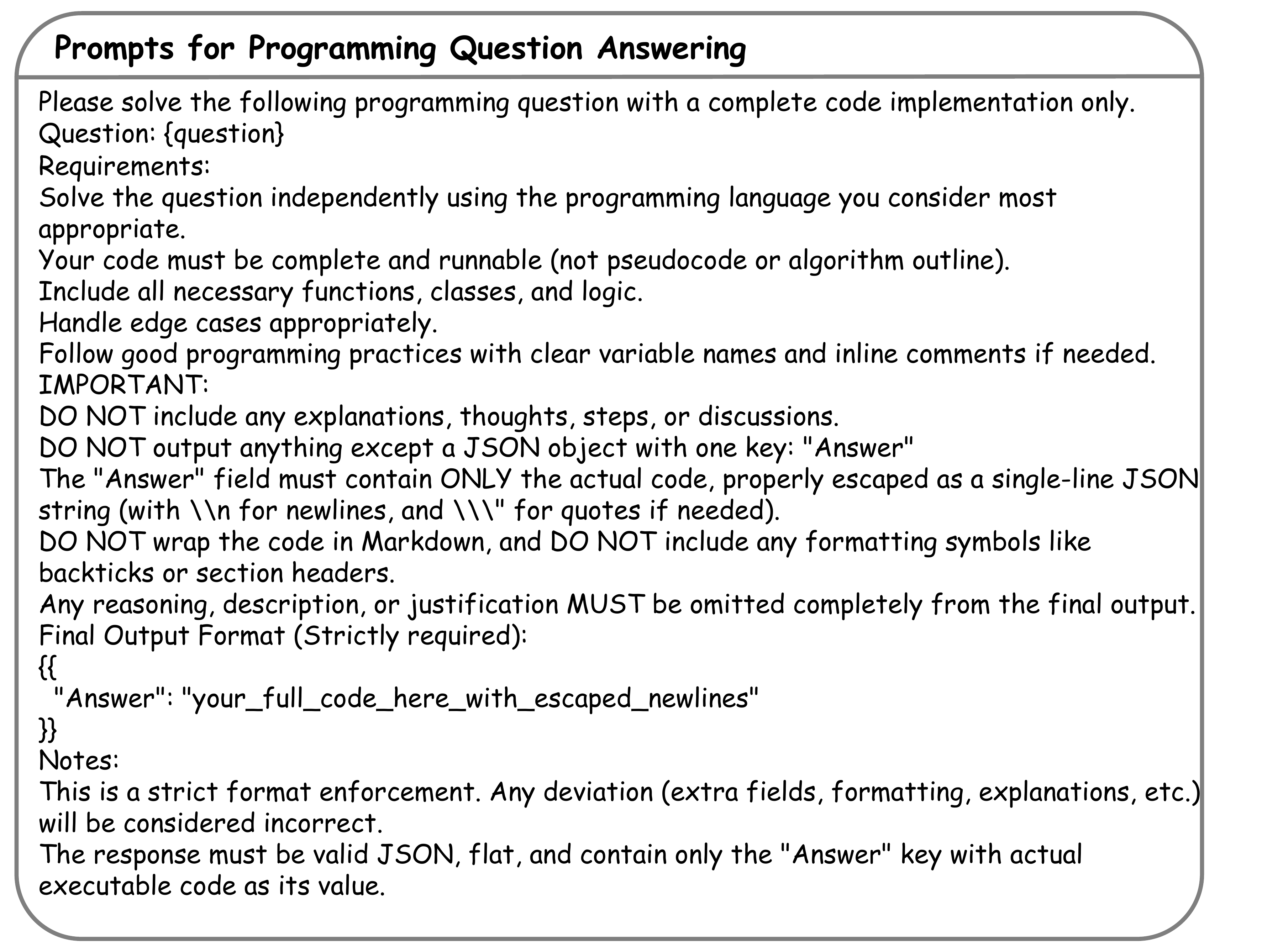}
\caption{Prompts for Programming Question Answering.}
\label{fig:prog_prompt_ans}
\end{figure*}

\begin{figure*}[!htbp]
\centering
\includegraphics[width=0.98\textwidth]{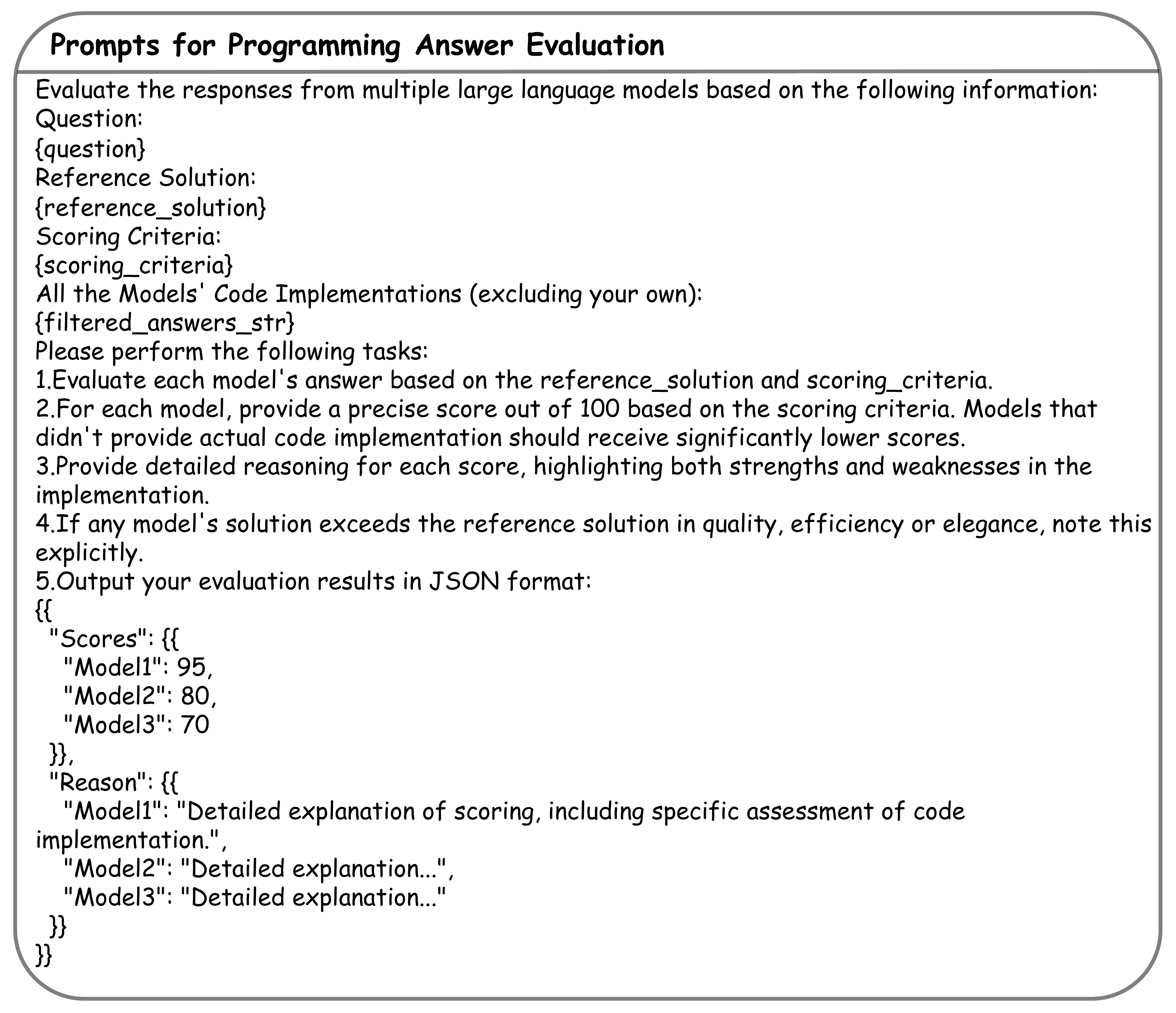}
\caption{Prompts for Programming Answer Evaluation.}
\label{fig:prog_prompt_eval}
\end{figure*}

The full prompts used for Programming include question generation, question answering, and answer evaluation.
The corresponding prompts are presented in Figures \ref{fig:prog_prompt_qgen}--\ref{fig:prog_prompt_eval}.

\end{document}